\documentclass[10pt,twocolumn,letterpaper]{article}

\usepackage{3dv}
\usepackage{times}
\usepackage{epsfig}
\usepackage{graphicx}
\usepackage{amsmath}
\usepackage{amssymb}
\usepackage{array}
\usepackage{comment}
\usepackage{booktabs}
\usepackage{multirow}
\usepackage[colorlinks,linkcolor=red,bookmarks=false]{hyperref} 
\usepackage{color}
\usepackage{cleveref}
\threedvfinalcopy % *** Uncomment this line for the final submission

\def\threedvPaperID{97} % *** Enter the 3DV Paper ID here

\ifthreedvfinal\fi
\begin{document}

%%%%%%%%% TITLE
\title{PanoDepth: A Two-Stage Approach for Monocular  Omnidirectional Depth Estimation}

\author{Yuyan Li\\
\small{University of Missouri}\\
{\tt\small yl235@umsystem.edu}
\and
Zhixin Yan\\
\small{BOSCH Research China}\\
{\tt\small zhixin.yan2@cn.bosch.com}
\and
Ye Duan\\
\small{University of Missouri}\\
{\tt\small duanye@umsystem.edu}
\and
Liu Ren\\
\small{BOSCH Research North America}\\
{\tt\small liu.ren@us.bosch.com}
\vspace{-0.5cm}
}

\maketitle
\thispagestyle{empty}

%%%%%%%%% ABSTRACT
\begin{abstract} Omnidirectional 3D information is essential for a wide range of applications such as Virtual Reality, Autonomous Driving, Robotics, etc. 
In this paper, we propose a novel, model-agnostic, two-stage pipeline for omnidirectional monocular depth estimation. Our proposed framework \textit{PanoDepth} takes one 360 image as input, produces one or more synthesized views in the first stage, and feeds the original image and the synthesized images into the subsequent stereo matching stage. 
In the second stage, we propose a differentiable Spherical Warping Layer to handle omnidirectional stereo geometry efficiently and effectively. By utilizing the explicit stereo-based geometric constraints in the stereo matching stage, \textit{PanoDepth} can generate dense high-quality depth. We conducted extensive experiments and ablation studies to evaluate \textit{PanoDepth} with both the full pipeline as well as the individual modules in each stage. Our results show that \textit{PanoDepth} outperforms the state-of-the-art approaches by a large margin for 360 monocular depth estimation. 
%Our code is available at 
%\url{https://github.com/yuyanli0831/PanoDepth_3dv}.
\end{abstract}
%%%%%%%%% BODY TEXT
\section{Introduction}
Omnidirectional 3D information is essential for a wide range of applications (e.g. Virtual Reality \cite{argyriou2020design}, augmented reality \cite{berning2013parnorama}, autonomous driving \cite{appiah2011obstacle}, and robotics \cite{pudics2015safe}). Quick and reliable omnidirectional data acquisition can facilitate many use cases, such as user interaction with the digital environment, robot navigation, and object detection for autonomous vehicles. Another relevant application is remote working/shopping/education \cite{Feurstein2018TowardsAI}, which has become ubiquitous due to the pandemic.  To obtain high-quality omnidirectional 3D information, devices such as omnidirectional LiDARs are widely used in autonomous driving and indoor 3D scans. However, LiDARs are either very expensive or can only produce sparse 3D scans. Compared with LiDARs, cameras are much cheaper and already frequently used for capturing the visual appearance of the scenes. The cost can be significantly reduced if high-quality omnidirectional 3D can be generated directly from camera images.

\begin{figure*}
    \centering
    \includegraphics[width=17cm]{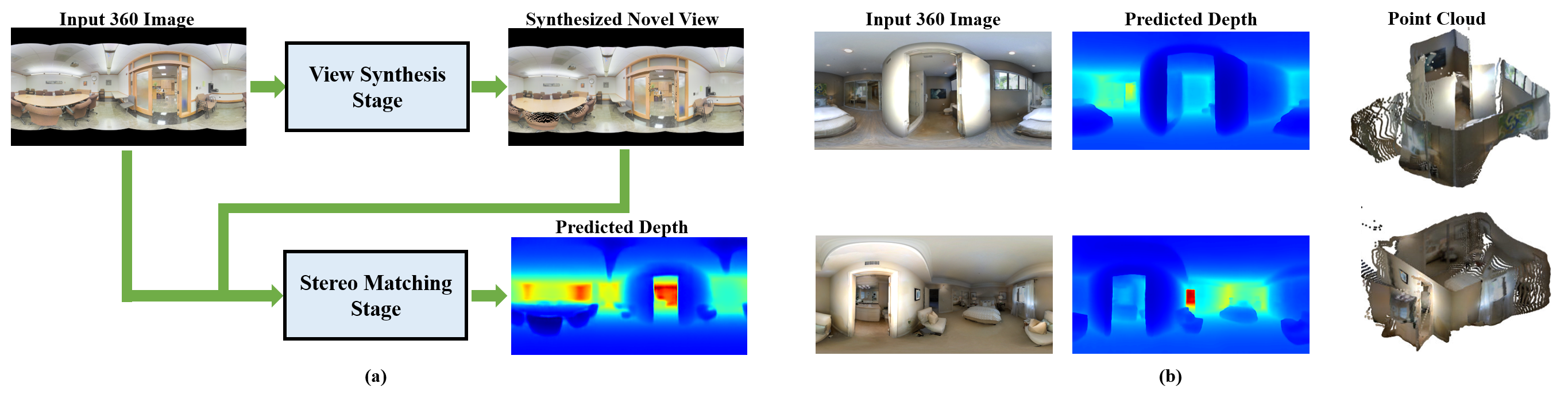}
    \caption{(a) Illustration of our \textit{PanoDepth} framework. \textit{PanoDepth} takes one 360 image as input to generate one or more novel views in the first view synthesis stage. The original and synthesized 360 images are then fed into the second multi-view stereo matching stage to predict final dense depth map. (b) Two examples (top and bottom row) of \textit{PanoDepth} on 360D dataset \cite{zioulis2018omnidepth} with 360 image (left) as input, and output depth (middle) and point cloud (right). }
    \label{fig:pipeline}
\end{figure*}

%Inferring dense depth from a monocular input is a challenging task in computer vision. 
Deep learning techniques, coupled with a growing accessibility of large-scale datasets, have largely improved the performance of many computer vision tasks including depth estimation \cite{zhao2020monocular}. Depth estimation often uses either a monocular input or a stereo pair. For the monocular methods, a common practice is to train a single network to map RGB pixel to real-value depth \cite{eigen2014depth, fu2018deep, hu2019revisiting, huynh2020guiding}, mostly by learning from various monocular cues such as shape, lighting, shading, object type, etc.  Stereo matching methods \cite{scharstein2002taxonomy,kendall2017end,chang2018pyramid}, on the other hand, learn the disparity by matching image patches from stereo pairs and later convert disparity to depth. Despite the significant improvement in monocular estimation methods, there is still a large gap between monocular and stereo depth accuracy
\cite{smolyanskiy2018importance}. 

To apply stereo matching for monocular depth estimation, Luo et al. \cite{luo2018single} proposed a two-stage pipeline that decomposes monocular depth estimation into two stages, view synthesis and stereo matching respectively. In their approach, a second view is first synthesized and fed together with the original view to the stereo matching stage to compute the disparity. Stereo matching can leverage more geometric constraints into the network training, thus reducing the demand for ground truth depth. More recent studies  \cite{tosi2019learning, watson2020learning} improved upon this idea and achieved promising performances. These two-stage methods \cite{luo2018single,tosi2019learning,watson2020learning}, are mainly designed for perspective images. In 360 domain, most of the recent studies \cite{zioulis2018omnidepth, su2017learning, eder2019mapped, wang2020bifuse, eder2019pano}
still follow the same single-stage monocular estimation procedure with adaptation to 360 \footnote{We use the terms 360, omnidirectional, equirectangular, spherical interchangeably in this paper} geometry.

In this paper, we propose a novel, model-agnostic, two-stage pipeline (see Figure \ref{fig:pipeline}) for solving
%that we successfully apply to solve 
the problem of 360 monocular depth estimation. Our proposed framework \textit{PanoDepth} takes one equirectangular projection (ERP) image as input, produces one or more synthesized views in the first stage, and feeds the original image and the synthesized images to the subsequent
stereo matching stage to predict the final depth map. 
In the stereo matching stage, we propose a novel %/omnidirectional stereo matching network with a 
differentiable Spherical Warping Layer to handle omnidirectional stereo geometry efficiently and effectively. 
We conducted extensive experiments and ablation studies to evaluate \textit{PanoDepth} with both the full pipeline and the individual networks in each stage on several public benchmark datasets.
Our results demonstrated that our model-agnostic approach \textit{PanoDepth} outperforms the one-stage method by a large margin despite the combinations of coarse estimation and stereo matching networks. Moreover, by adjusting these networks, \textit{PanoDepth} can be adapted to the target computation constraints and performance requirements. 
%Our results show that \textit{PanoDepth} is flexible and model-agnostic. With various types of coarse estimation network and stereo matching methods, the two-stage \textit{PanoDepth} outperforms the one-stage-only methods.
%Our results show that \textit{PanoDepth} drastically improves the depth quality with a monocular input.
%To promote future research in 360 depth estimation we will also release a new benchmark dataset \textit{Pano3D}. 

Our contributions can be summarized as follows:
\begin{itemize}
%\item We propose a novel, \ye{model agnostic,} two-stage end-to-end framework \textit{PanoDepth}, consisting of view synthesis and stereo matching. %\textit{PanoDepth} fully exploits the synthesized 360 views and stereo constraints to improve omnidirectional depth quality and outperforms the state-of-the-art monocular omnidirectional depth estimation approaches by a large margin.

\item We propose a novel, model-agnostic, two-stage framework \textit{PanoDepth}, including view synthesis and stereo matching,
to fully exploit the synthesized 360 views and spherical stereo constraints.

\item \textit{PanoDepth} outperforms the state-of-the-art monocular omnidirectional depth estimation approaches by a large margin.

\item We propose a novel differentiable Spherical Warping Layer (SWL) which adapts regular stereo matching networks to 360 stereo geometry, and enables advanced features such as multi-view stereo and cascade mechanism for stereo performance boost.
%\yuyan{spherical warping layer and the stereo matching methods  effectively adapts to 360 stereo geometry and achieves state-of-the-art performance}.
%\item A new benchmark \textit{Pano3D} will be released to promote future omnidirectional depth estimation research. 
\end{itemize}

\section{Motivation}
\label{sec:motivation}
In this section, we explain the motivation of formulating the 360 monocular depth estimation problem as two separate stages, namely, a view synthesis stage based on coarse depth estimation, and a multi-view stereo matching stage for final depth output.

\subsection{Why Two-Stage?}
One main advantage of monocular depth estimation is its potential in dramatically reducing the hardware cost for 3D depth acquisition. Motivated by this, many studies have been proposed to solve this problem.
The basic idea of supervised monocular estimation methods is to train a network that directly learns the mapping from the input RGB pixels to the real-value output depth in a single stage. For example, Laina et al. \cite{laina2016deeper} proposed FCRN which uses ResNet-50 \cite{he2016deep} as backbone, followed by multiple up-projection modules. Hu et al. \cite{hu2019revisiting} leveraged SENet-154 \cite{hu2018senet} as encoder together with multi-scale fusion module. 

On the other hand, deep learning based stereo matching networks \cite{chang2018pyramid, kendall2017end, gu2019cascade} utilize the stereo constraints to improve efficiency and the output depth quality. These methods simulate the traditional stereo matching process by learning and optimizing the matching cost across the input image pairs in a deterministic manner. 
%Depth estimation can then be formulated as a image patch correspondence matching problem \cite{eigen2014depth}. 
Unlike monocular depth methods which directly map RGB into depth by considering all the monocular cues, stereo matching methods focus on estimating disparity by developing image patch correspondence \cite{eigen2014depth}. Given the predefined baseline and 1D search space along the epipolar line for image patch matching \cite{hartley2003multiple}, stereo matching produces more accurate depth maps in comparison with monocular methods in general \cite{smolyanskiy2018importance}. 

A typical stereo matching network requires at least two images as input, which is not directly applicable for a monocular input setting.
However, if one or more novel views can be synthesized with high quality, these additional views can be utilized to train a stereo matching network. Luo et al. \cite{luo2018single} first proposed the two-stage pipeline for perspective images where a novel right view is synthesized at the first stage and paired with the original view to the second stereo stage.
%In their approach, they estimate a novel right view with view synthesis network from the original left view, then process both of the synthesized view and the original view to the stereo matching network. 
Later, two-stage approaches \cite{tosi2019learning} mostly followed this work to generate a coarse disparity/depth, and to synthesize novel views via image warping or Depth-Image-Based Rendering (DIBR). Taking the original and the synthesized views as input, the final depth generated from the later stereo matching stage of these approaches shows a significant improvement over the one-stage counterparts. 

\subsection{Can we successfully synthesize novel view 360 images?}
Recent two-stage approaches \cite{luo2018single, tosi2019learning} have shown promising capabilities in improving depth  quality on perspective images. However, it remains unclear whether the two-stage approach will be applicable for 360 images, as there are many fundamental differences between the perspective images and 360 images, such as camera projection model, image distortion, and field of view (FoV).
%An illustration of equirectangular projection model is shown in Figure \ref{fig:projection_model}.% Furthermore, very limited studies have explored view synthesis as well as learning based stereo matching in 360 domain. In this paper, as the first of its kind, we investigate the use of this two-stage pipeline for 360 monocular depth estimation.

The difference in camera projection model, equirectangular projection (see Figure \ref{fig:projection_model}(a)) vs. perspective projection, can be resolved by integrating spherical geometry into the disparity calculation and cost volume fusion procedure. In this paper, we propose a novel Spherical Warping Layer specifically designed for spherical geometry as a solution (Section \ref{subsec:MVSNet}). Moreover, the distortion issue can be addressed by applying distortion-aware convolutions \cite{eder2019mapped, eder2019convolutions, su2019kernel, su2017learning, coors2018spherenet, zhao2018distortion, chen2021distortion}. Hence, in this section we will mainly discuss the difference in FoV settings. Comparing with perspective images, 360 images have much larger FoVs ($360^{\circ}$ horizontally, $180^{\circ}$ vertically). A 360 image encodes almost every piece of visual information of the scene except occluded areas, while perspective images suffer from information loss near the image boundaries in addition to occlusions. This could be a great advantage for novel view synthesis of 360 images. 

To validate this observation, we conducted various experiments regarding the correlations between image FoV, baseline and synthesized view quality (more details in the Appendix). Our experiment confirms that i) with greater FoV, the synthesized views are less sensitive to large baselines, and ii) synthesized 360 images have the least error and artifacts. According to Gallup et al. \cite{gallup2008variable}, depth error of stereo matching comes from both the disparity error (proportional) and the baseline (inversely proportional). Thus, with higher quality synthesized novel views and larger baselines, we expect less error in the final depth output from stereo matching. This also indicates that the two-stage pipeline is well-suited for 360 monocular depth estimation.

%------------------------------------------------------------------------
\begin{figure*}[hbt!]
\centering
  \includegraphics[width=17cm]{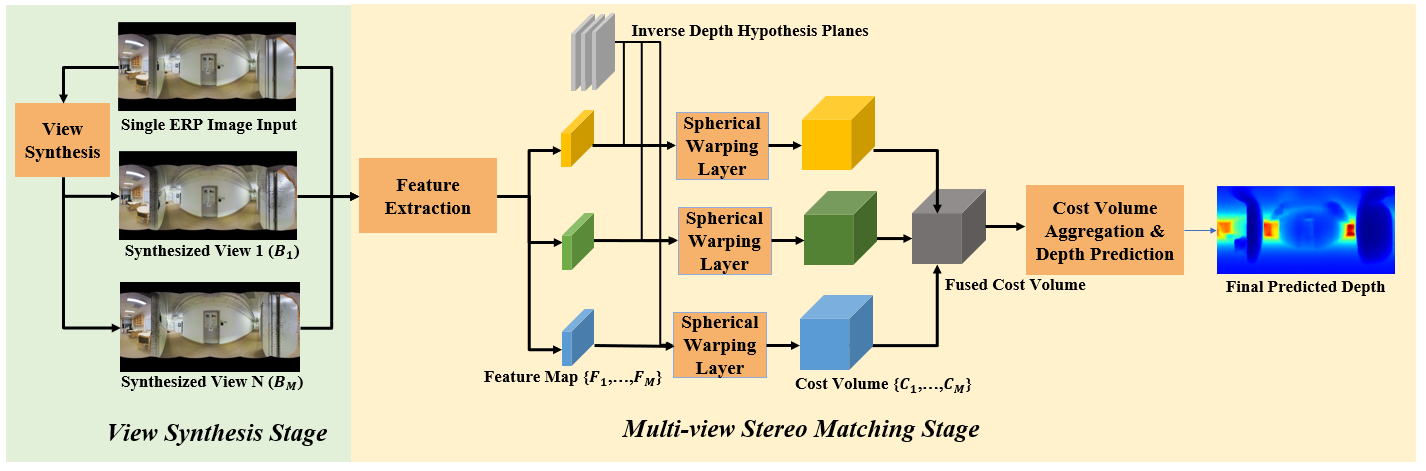}
  \caption{Illustration of our \textit{PanoDepth} framework which consists of a view synthesis stage and a multi-view stereo matching stage. In the view synthesis stage, a total of $M$ synthesized views are generated.  In the multi-view stereo matching stage, the synthesized 360 views together with the original input view, are sent to the multi-view stereo matching network to produce the final depth estimation. To better adapt to 360 stereo geometry, we directly sample hypothesis plane on the inverse depth, and use a Spherical Warping Layer (SWL) to warp reference views to the target view (Section \ref{subsec:MVSNet}).}
  \label{fig:stereo_matching_network}
\end{figure*}

\section{Related Work}
\label{sec:related}

\subsection{Monocular Depth Estimation}
%Researchers have been studying the estimation of depth from monocular perspective images for more than a decade
Monocular depth  estimation
\cite{saxena2006learning} has seen significant improvements \cite{laina2016deeper, hu2019revisiting} since the first adoption of  deep learning by Eigen et al.  \cite{eigen2014depth}. %proposed the first monocular depth estimation network. 
%Recent approaches have been looking at the improvement of depth map quality with deeper networks \cite{laina2016deeper, hu2019revisiting}. 
To further improve the performance, researchers  explored many strategies such as multi-task learning with normal estimation \cite{qi2018geonet} and semantic segmentation \cite{eigen2015predicting, jiao2018look} along with depth estimation, incorporating CRF \cite{cao2017estimating, mahmood2018deep}, integrating attention modules \cite{huynh2020guiding, li2018deep}, utilizing planar constraints \cite{liu2018planenet, huynh2020guiding}, conducting unsupervised learning using constrainsts such as left-right consistency \cite{godard2017unsupervised, bian2019unsupervised, payen2018eliminating}, as well as two-stage approaches where stereo constraints are leveraged \cite{luo2018single,tosi2019learning,watson2020learning}.

As 360 cameras become more affordable, researchers start to explore the possibility to use 360 images for depth estimation \cite{eder2019mapped, Tateno_2018_ECCV, cheng2020omnidirectional, zioulis2018omnidepth, wang2020360sd, zioulis2019spherical, zeng2020joint}. %including monocular depth estimation. Most of these methods utilize the equirectangular projection(ERP) representation, as it does not has the boundary discontinuity artifact like the alternative Cube-map representation. However as a result of the camera projection procedure, ERP does suffer image distortion which can be resolved using either spherical kernels \cite{coors2018spherenet, su2019kernel, eder2019mapped, su2017learning}or distortion-aware convolutions \cite{Tateno_2018_ECCV}. 
To advance the performance of 360 monocular depth estimation, Eder et al. \cite{eder2019pano} proposed joint training of surface normal, boundary, and depth. Zeng et al. \cite{zeng2020joint} trained a network which combines 3D layout and depth. Jin et al. \cite{Jin_2020_CVPR} took advantage of the correlation between depth map and geometric structure of 360 indoor images. Cheng et al. \cite{cheng2020omnidirectional} proposed a low-cost sensing system which combines an omnidirectional camera with a calibrated projective depth camera. The 360 image and the limited FoV depth are used together as input to a CNN. Meanwhile, distortion-aware convolution filters \cite{Tateno_2018_ECCV,zhao2018distortion, eder2019mapped,chen2021distortion} are designed to handle spherical geometric distortion.

%Different from the previous works, we propose to follow the steps of \cite{luo2018single} and use a two-stage pipeline on 360 images in this paper. Our framework first synthesizes novel 360 views which are sufficient for stereo matching network training, and thus can enforce strong and explicit geometric constraints into the network. 
%Experiments show our framework outperform current state-of-the-art approaches by a large margin on multiple benchmark datasets.

\subsection{Multi-View Stereo Matching}
Besides monocular depth estimation, Multi-View Stereo (MVS) is another group of methods for predicting depth. Given a set of images with known camera poses, MVS approaches \cite{huang2018deepmvs, yao2018mvsnet} can produce highly accurate depth estimates with multi-view geometric constraints.
One example is MVSNet \cite{yao2018mvsnet}, in which the variance-based cost volume is presented to fuse multiple features maps from source images into one unified cost volume. Stereo matching can be treated as a special case of MVS. 
%Besides monocular depth estimation, stereo matching is the most explored area for inferring 3D information. 
Conventionally, stereo-based depth estimation methods \cite{scharstein2002taxonomy, hirschmuller2005accurate} relied on matching pixels across stereo images. %The success of deep learning along with the accessibility of large training dataset has gradually replaced the traditional methods and achieved excellent performance. 
Many recent stereo matching approaches \cite{kendall2017end,chang2018pyramid} leveraged CNNs for feature extraction, cost matching, and aggregation. For example, PSMNet \cite{chang2018pyramid} incorporated spatial pyramid pooling (SPP) module and multi-scale 3D hourglass modules to further boost the performance. To improve the efficiency and accelerate training on high-resolution images, Gu et al. \cite{gu2019cascade} presented a cascade cost volume design to gradually retrieve finer hypothesis plane ranging over multiple steps.

In 360 stereo domain, SweepNet \cite{won2019sweepnet} and
OmniMVS \cite{won2019omnimvs} estimated depth from multiple wide-baseline fisheye cameras. 
Another recent work is 360SD-Net \cite{wang2020360sd} which predicted disparity/depth from a pair of ERP images that are taken by a top-bottom camera pair. They \cite{wang2020360sd} incorporated polar angles to solve distortions and proposed learnable shifting filters to adjust the step size in disparity cost volume construction. However, the learnable shifting filters create extra overhead during training. 
In this paper, we propose a closed-form solution, Spherical Warping Layer (SWL), that does not require additional training overhead. Our experiments (Table \ref{tab:ablation_stereo}) show that SWL can significantly improve the performance of 360 stereo matching. %and compares favorably over existing methods with no SWL \cite{chang2018pyramid,wang2020360sd}.}

%In our approach, we demonstrate that the step size can be automatically computed with a closed-form solution using a Spherical Warping Layer instead of the learnable filters. Experiments show that our stereo matching outperforms existing 360 stereo matching methods such as PSMNet \cite{chang2018pyramid} and 360SD-Net \cite{wang2020360sd}.

\section{Approach}
\label{sec:approach}

We propose an end-to-end framework, \textit{PanoDepth}, that takes a single ERP image as input and produces a high-quality omnidirectional depth map. 
\textit{PanoDepth}  consists of two stages: i) a view synthesis stage that conducts coarse depth estimation followed by a differentiable DIBR module for novel view synthesis, and ii) a stereo matching stage with a customized Spherical Warping Layer for efficient and high-quality 360 depth estimation. A full framework of \textit{PanoDepth} is illustrated in Figure \ref{fig:pipeline}. 

\subsection{View Synthesis Stage}
\label{subsec:view_synthesis}
%\zhixin{In previous two-stage approaches (e.g. \cite{luo2018single,zhou2018stereo}), a coarse depth map is usually estimated first and then combined with DIBR module for novel view synthesis. Based on our empirical observations (see the Appendix for details), such procedure also works well for 360 novel view synthesis with different configurations of coarse depth estimation networks. }

To synthesize high-quality novel views, the coarse depth map is usually estimated first followed by a Depth-Image-Based Rendering (DIBR) module \cite{luo2018single,zhou2018stereo}. % are generally sufficient. %based on our empirical observations (see the Appendix for details). 
Based on our empirical observations (see the Appendix for details), such a procedure also works well for 360 novel view synthesis with different configurations of coarse depth estimation networks. Considering both performance and computation cost, in this paper we suggest using a lightweight network: CoordNet \cite{zioulis2019spherical} for doing the task. CoordNet utilizes coordinate convolution \cite{liu2018intriguing} to enforce 360 awareness. We 
append an atrous spatial pyramid pooling module (ASPP) \cite{chen2017rethinking} to the end of the encoder to better aggregate  multi-scale context information. Note that \textit{PanoDepth} is model-agnostic, thus any depth estimation network can be used here to fulfill specific requirement.
The estimated coarse depth map and the original ERP image are then used to render multiple synthesized views of predefined baselines via a differentiable DIBR operation \cite{tulsiani2018layer}. In this paper, we choose to use vertical baselines instead of horizontal ones. The analysis of this choice can be found in the Appendix.

\subsection{Stereo Matching Stage}
\label{subsec:MVSNet}
The second stage of our \textit{PanoDepth} framework is stereo matching. Again, as \textit{PanoDepth} is model-agnostic, any stereo matching network can be plugged in here. Experimental results that show the performance of different stereo matching network settings is discussed in Section \ref{subsec:ablation}. 

The stereo matching network we used in this paper follows a similar pipeline as PSMNet \cite{chang2018pyramid}, with several key modifications.
The network consists of five main modules:
feature extraction, spherical warping layer, cost volume construction, cost aggregation, and depth prediction. Comparing with the original PSMNet \cite{chang2018pyramid}, our unique contribution is the Spherical Warping Layer (SWL) which is specifically designed for the 360 stereo geometry.

\textbf{Feature Extraction}
After the view synthesis stage, the input image along with all the $M$ synthesized novel views will be passed to a weight-sharing neural network to extract features. We use the same layer setting and keep the SPP module as the original PSMNet \cite{chang2018pyramid}.

\textbf{Spherical Warping Layer (SWL)} %After view synthesis stage, the input image along with all the $M$ synthesized novel views will be passed to a weight-sharing neural network to extract features. 
The extracted feature maps of all views are then used to build a cost volume at multiple depth hypothesis planes for cost matching.
An essential step of cost volume construction is to determine the coordinate mapping, which is reflected as disparity, that warps reference view to the target view.
Unlike perspective images where the disparity is proportional to the inverse depth \cite{pollefeys1999simple, chang2018pyramid, valentin2018depth}, disparity of 360 stereo pairs is related to both inverse depth and spherical latitudes. 
Our SWL performs direct depth sampling instead of disparity sampling which is commonly used in perspective image pair stereo matching. Compared with the learnable shifting filters proposed in \cite{wang2020360sd}, our SWL makes the disparity computation adjustable to the pixel latitudinal value, without introducing additional computation overhead in training. 
%Moreover, as a closed-form solution, SWL naturally adapts to various baselines and inverse depth sampling range, which enables the adaptation of multi-view stereo and cascade mechanism in 360 stereo matching.
Moreover, SWL directly samples on the absolute depth domain, thereby enabling horizontal 360 stereo (which has both horizontal and vertical disparity, more details in the Appendix), the adaptation of 360 multi-view, and the usage of cascade mechanism.

\begin{figure}[t]
    \centering
    \includegraphics[width=8.5cm]{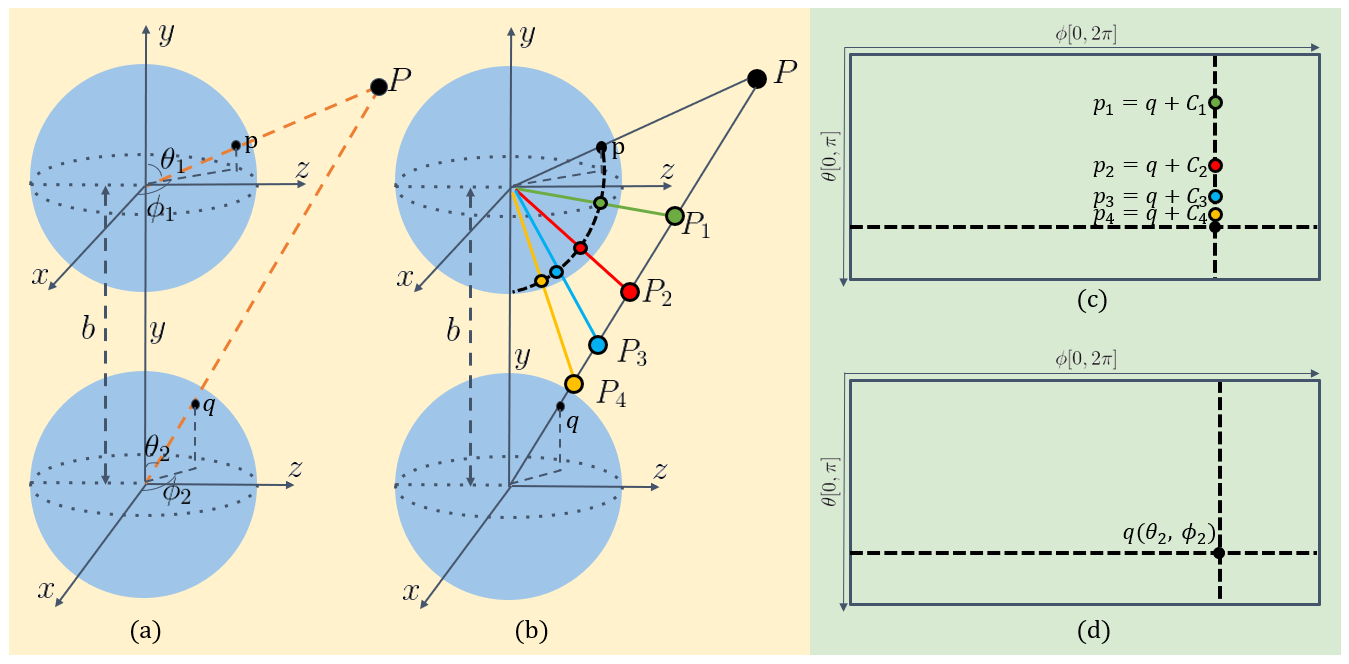}
    \caption{Visualization of our spherical warping method. (a) Vertical 360 stereo model. $b$ is the baseline displacement of two cameras. $P$ is a real-world point. The projection of $P$ on two camera space is represented as $p(\phi_1, \theta_1)$ and $q(\phi_2, \theta_2)$.
    (b) Spherical epipolar geometry. $P_i$ is the points sampled at different depths. (c) Projection of sampled inverse depth on the ERP image. $p_i$ is the projection of $P$ on the top view at sampled points $P_i$. $C_i$ is the vertical disparity, it equals to $C_y$ in Equ (\ref{eq:sw}). (d) Projection on the reference image. $q$ is the projection of $P$ at bottom view. }
    \label{fig:projection_model}
\end{figure}

Figure \ref{fig:projection_model} shows an illustration of the Spherical Warping Layer. We first  sample the inverse depth  to cover the whole depth range:
\begin{equation}
    \frac{1}{d_j} = \frac{1}{d_{max}} + (\frac{1}{d_{min}}-\frac{1}{d_{max}})\frac{v\times j}{D-1}, j=0,1,..., D-1
    \label{eq:hypothesis}
\end{equation}
where $D$ is the total number of hypothesis planes, $d_j$ is the $j^{th}$ depth plane, $d_{min}$ and $d_{max}$ are the minimum and maximum of the depth image, $v$ is the plane interval. 
The Spherical Warping Layer then transforms depth hypothesis $d_j$ to displacement in spherical domain $C_j$, to map pixels from the reference synthesized view to the target view. The displacement $C_j$ is defined as:
\begin{equation}
   C_{x,j}=0, \; C_{y,j} = \frac{cos(\theta)\times b}{d_j} \times \frac{H_f}{\pi} 
   \label{eq:sw}
\end{equation}
where $\theta$ refers to the pixel-wise latitudinal values, $b$ represents the baseline, and $H_f$ is the height of the feature map. %Compared with the learnable shifting filters proposed in \cite{wang2020360sd}, our Spherical Warping Layer has no computation overhead in training since it is a closed-form solution. 

\textbf{Cost Volume Construction} 
The SWL transforms reference view feature maps into the target view domain at the individual hypothesis plane, and then a total of $M+1$ feature volumes are generated.
The variance-based cost volume formation method from MVSNet \cite{yao2018mvsnet} is used for the fusion of these %$M+1$ 
feature volumes into a compact one. 
Moreover, we adopt a cascade design from Gu et al. \cite{gu2019cascade} to further improve the final depth quality. 
Specifically, at level $l(l>1)$, $d_{min}$ and $d_{max}$ is recalculated based on the prediction of level $l-1$, then the new depth range and the new number of planes $D_l$ is used to determine the new intervals. Depth hypothesis for level $l$ is then updated using equ (\ref{eq:hypothesis}). 
The corresponding displacements are calculated via the same spherical coordinate mapping procedure.  
%\zhixin{Other more advanced multi-view stereo and cascade design approaches \cite{huang2018deepmvs, yu2020fastmvsnet, shen2020msmd} can also be used here to further improve the performance. However, since the main scope of this paper is about the use two-stage pipeline in 360 monocular depth estimation, we believe the current setup can already demonstrate the advantage of our two-stage approach.}

\textbf{Cost Aggregation and Depth Prediction}
After the construction of the cost volume, multi-scale 3D CNN is used  to aggregate different levels of spatial context information through the hourglass-shape encoding and decoding network. It has been shown this kind of cost aggregation module helps to regularize noises in ambiguous regions caused by occlusions, textureless surfaces, and to improve final prediction quality \cite{kendall2017end, chang2018pyramid, guo2019group}. Finally, we regress the depth value at each level $l$:
\begin{align}
   &\frac{1}{d_{pred, l}} = \frac{1}{d_{min, l}} + (\frac{1}{d_{min, l}}-\frac{1}{d_{max, l}})\frac{k_l}{D_l-1} \\
   &k_l = \sum^{D_l-1}_{j=0} \sigma(p_j) \times (v_{l}\times j)
\label{eq:prediction}
\end{align}
where $k_l$ is the sum of each plane level weighted by its normalized probability, $\sigma(\cdot)$ represents softmax function, $p_j$ denotes the probability of $j^{th}$ plane value, $v_{l}$ is the interval for level $l$.

\subsection{Loss Function}
\textit{PanoDepth}  is trained in an end-to-end fashion, supervision is applied on both stages. The final loss function is defined as follows,
\begin{equation}
    \mathcal{L}_{total} = \omega_1\mathcal{L}_{coarse} + \omega_2\mathcal{L}_{stereo}
\end{equation}
where $\omega_1$ and $\omega_2$ are the weights of coarse depth estimation loss and stereo matching loss respectively. For the optimization on the first stage coarse estimation, we use inverse Huber (berHu) loss as proposed in \cite{laina2016deeper}:
\begin{equation}
   \mathcal{L}_{coarse} = \frac{1}{\Omega} \sum_{i\in \Omega} \mathcal{L}_{berHu}(d_{i},\;\;d_{i}^*) \\
\end{equation}
where $\Omega$ is a binary mask that is used to mask out missing regions (pixels that have depth values smaller than $d_{min}$ or greater than $d_{max}$), $d_i$ and $d_i^*$ are the ground truth and the predicted depth value of a valid pixel $i$ respectively. For stereo matching, we calculate berHu loss \cite{laina2016deeper} on all outputs from each level $l$ and then compute the weighted summation. The stereo matching loss is defined as:
\begin{equation}
    \mathcal{L}_{stereo} = \frac{1}{\Omega}\sum_{i\in \Omega} \sum_{l=1}^{N} \lambda_l \mathcal{L}_{berHu}(d_i,\;\;d_{i}^*)
\end{equation}
where $\lambda_l$ is the level $l$ stereo loss weight.

\section{Experiments}
\label{sec:exp}
\subsection{Datasets}
\label{subsec:dataset}
We train and evaluate our network on three panorama RGBD benchmark datasets including, Stanford2D3D \cite{armeni2017joint}, 360D \cite{zioulis2018omnidepth} and the omnidirectional stereo dataset \cite{zioulis2019spherical}. 

\textbf{Stanford2D3D} Stanford2D3D \cite{armeni2017joint} dataset consists of 1413 real-world panorama images from six large-scale indoor areas. We follow the official train-test split which uses the fifth area for testing, and other areas for training. We resize the images to $256 \times 512$ to reduce computation time.

\textbf{360D} 360D \cite{zioulis2018omnidepth} is a RGBD panorama benchmark provided by Zioulis et al. \cite{zioulis2018omnidepth}. It is composed of two other synthetic datasets (SunCG and SceneNet), and two real-world datasets (Stanford2D3D and Matterport3D). There are 35,977 panorama RGBD images in the 360D that are rendered from the aforementioned four datasets. We again follow the default train-test splits. 

\textbf{Omnidirectional Stereo Dataset}. The omnidirectional stereo dataset \cite{zioulis2019spherical} consists of 7964 stereo pairs of panorama RGBD images rendered from two real-world datasets, Matterport3D \cite{Matterport3D} and Stanford2D3D \cite{armeni2017joint}. We use the train-test split that removes 3 complete buildings from Matterport3D \cite{Matterport3D} and 1 complete area from Stanford2D3D \cite{armeni2017joint} for test. Each set of data consists of left-down, right, and up view 360 RGBD images in a triangular fashion with size $256\times 512$. We only use images from the left-down view in our single view depth estimation experiments. Up-down stereo pairs are only used for the ablation study of the stereo matching network. 

%\textbf{Pano3D} 
To investigate the impact of baselines and FoV on the quality of view synthesis, we create a new dataset that is rendered from the mesh of Stanford2D3D \cite{armeni2017joint}. More details of this new dataset as well as the experiments with various baseline configurations are included in the Appendix.

\subsection{Implementation Details and Metrics}
\label{subsec:impl}
For parameter settings, we use a default of $N=2$ levels, with $D_1=48$ and $D_2=24$ hypothesis planes respectively. The minimum and maximum depth $d_{min}$ and $d_{max}$ for the first level is set to be $0.2m$ and $8m$. We use a default of $M=3$ synthesized views rendered at vertical baseline placements $-0.24m, +0.24m, +0.4m$. The loss weights $\omega_1$ and $\omega_2$ are set to $1$ and $0.02$.
We train our framework from scratch using Adam optimizer ($\beta_1=0.9, \beta_2=0.999$) with a batch size of 8. Initial learning rates for the first and the second stage are set to 0.0002 and 0.0005. We separately train the coarse network for 10 epochs and then train the entire framework end-to-end for 200 epochs. 
Both of the learning rates decay by a factor of 0.5 every 30 epochs. 
Performances are evaluated based on commonly used depth quality measures \cite{eigen2014depth}: absolute relative error (Abs Rel), square relative error (Sq Rel), linear root mean square error(RMSE) and its natural log scale (RMSE log) and inlier ratios ($\delta_i < 1.25^i$, $i \in \{1, 2, 3\}$).

\subsection{Overall Performance Comparison with the State-of-the-art Algorithms}
\label{subsec:stoa}
\begin{figure*}[hbt!]
    \centering
    \includegraphics[width=17cm]{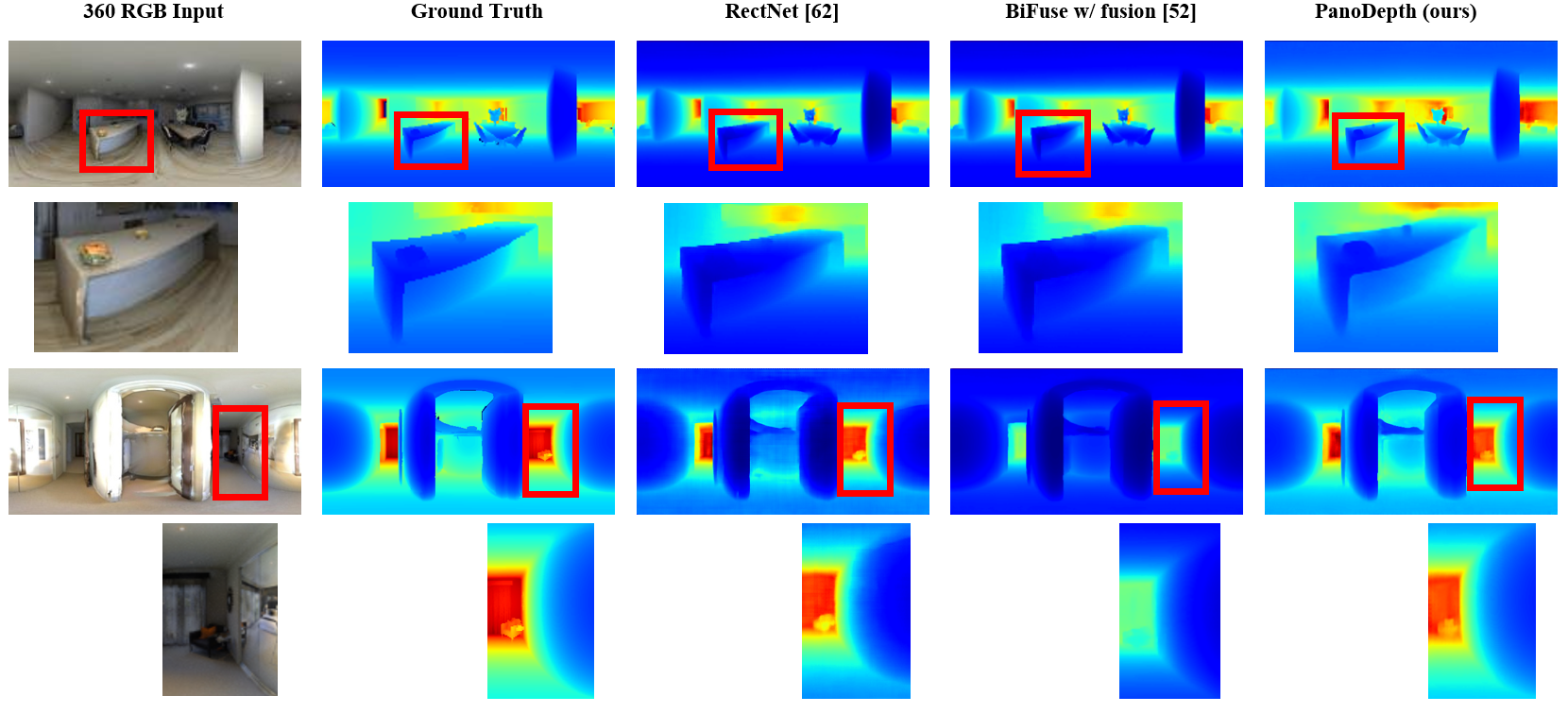}
    \caption{A qualitative comparison between RectNet \cite{zioulis2018omnidepth} (3rd column), BiFuse \cite{wang2020bifuse} (4th column), and  our method (5th column) on 360D \cite{zioulis2018omnidepth}. We highlight and zoom in some areas that distinguish the performance of three methods. We can see that our \textit{PanoDepth} is able to produce sharp edges, predict depth range accurately, and recover surface detail.}
    \label{fig:result}
\end{figure*}

Table \ref{tab:compare1} lists quantitative comparison between \textit{PanoDepth}  and other state-of-the-art omnidirectional monocular depth estimation methods \cite{laina2016deeper,zioulis2018omnidepth,wang2020bifuse,cheng2020omnidirectional} on both Stanford2D3D \cite{armeni2017joint} and 360D \cite{zioulis2018omnidepth} datasets. 
%We copy the results from \cite{wang2020bifuse} and compare with our method.
As shown in Table \ref{tab:compare1}, our method is able to reduce Abs Rel error by 19.60\% on Stanford2D3D and 25.85\% on 360D compare to the current leading 360 monocular depth estimation approach BiFuse \cite{wang2020bifuse}. Note that BiFuse \cite{wang2020bifuse} uses a network architecture with more than 200M parameters and has a large computation complexity for sharing information between CubeMap \cite{monroy2018salnet360} and ERP formats. The framework with distortion-aware module proposed in \cite{chen2021distortion} outperforms ours but it has more than 60M parameters.
Our framework has only around 16M parameters with a smaller computation overhead. Comparing the performance with ODE-CNN \cite{cheng2020omnidirectional} on 360D, our approach achieves comparable results while ODE-CNN requires additional depth sensor input. 
Figure \ref{fig:result} shows the qualitative comparison with the state-of-the-art approaches.  As we can see, our method generates high-quality depth with a detailed surface, sharp edges, and precise range. %\yuyan{It is worth to mention that our two-stage framework is agnostic, we can replace the coarse depth estimation with any monocular method in the first stage, or plug in other stereo matching networks in the second stage.The ablation studies for different variations of coarse depth estimation and stereo matching networks can be found in Appendix.}
%{In the appendix, we show that the performance do not degenerate much even using the one-level stereo network.}
%best results comparing  From the qualitative results, we can see that Bifuse and our approach have qualitatively comparable results. %More of the qualitative comparisons with Bifuse can be found Section in \ref{subsec:generalization}.

\begin{table*}[hbt!]
\begin{center}
\begin{footnotesize}
\begin{tabular}{m{7em} | m{16em} | m{4em}  m{4em}  m{3em}  m{3em}  m{3em} } 
\hline 
Datasets & Methods  & Abs Rel$\downarrow$ & RMSE$\downarrow$ & $\delta_1\uparrow$ & $\delta_2\uparrow$ & $\delta_3\uparrow$ \\
\hline \hline
\multirow{6}{*}{Stanford2D3D \cite{armeni2017joint}}  & FCRN \cite{laina2016deeper}  & 0.1837 & 0.5774  & 0.7230 & 0.9207 & 0.9731\\
 & RectNet \cite{zioulis2018omnidepth}   & 0.1409  & 0.4568 & 0.8326 & 0.9518 & 0.9822 \\
 & BiFuse with fusion \cite{wang2020bifuse} & 0.1209 & 0.4142 & 0.8660 & 0.9580 & 0.9860\\
 & Joint wth layout and semantics \cite{zeng2020joint} & 0.0680 & 0.2640 & 0.9540 & 0.9920 & 0.9980\\
\cline{2-7}
 & \textbf{PanoDepth(Ours)} & 0.0972 & 0.3747  & 0.9001 & 0.9701 & 0.9900\\
\hline
\hline
\multirow{7}{*}{360D \cite{zioulis2018omnidepth}} & FCRN  \cite{laina2016deeper} & 0.0699  & 0.2833  & 0.9532 & 0.9905 & 0.9966\\
 & RectNet \cite{zioulis2018omnidepth}  & 0.0702 & 0.2911 & 0.9574 & 0.9933 & 0.9979\\
 & Mapped Convolution \cite{eder2019mapped} & 0.0965 & 0.2966 & 0.9068 & 0.9854 & 0.9967\\
 & Distortion-aware \cite{chen2021distortion} & 0.0406 & 0.1769 & 0.9865 & 0.9966 & 0.9987 \\
 & BiFuse with fusion \cite{wang2020bifuse} & 0.0615 & 0.2440 & 0.9699 & 0.9927 & 0.9969\\
 & ODE-CNN \cite{cheng2020omnidirectional} & 0.0467  & 0.1728  & 0.9814 & 0.9967 & 0.9989\\

\cline{2-7}
 & \textbf{PanoDepth(Ours)} & 0.0456 & 0.1955 & 0.9830 & 0.9957 & 0.9984\\
\hline
\end{tabular}
\end{footnotesize}
\caption{A quantitative comparison with the state-of-the-art approaches on Stanford2D3D \cite{armeni2017joint} dataset and 360D \cite{zioulis2018omnidepth} dataset ($\downarrow$ represents lower the better, $\uparrow$ represents higher the better). We report the results based on the original papers \cite{zioulis2018omnidepth, wang2020bifuse, cheng2020omnidirectional} using the same evaluation metrics.
Note that ODE-CNN \cite{cheng2020omnidirectional} requires additional depth sensor input besides the 360 image used by other methods listed in the table. Additional supervision signals including layout and semantics are used in \cite{zeng2020joint}. 
For our \textit{PanoDepth}, we use the default stereo network setting with three synthesized views and a two-level cascade design. }
\label{tab:compare1}
\end{center}
\end{table*}

\subsection{Ablation Studies}
\label{subsec:ablation}
%In this section, we report the ablation studies of our \textit{PanoDepth} framework. 

\begin{table*}[hbt!]
\begin{center}
\begin{footnotesize}
\begin{tabular}{ m{22em} m{4em} m{4em}  m{4em}  m{5em}  m{3em}  m{3em}  m{3em} } 
\hline
Methods & Abs Rel$\downarrow$ & Sq Rel$\downarrow$ & RMSE$\downarrow$ & RMSElog$\downarrow$ & $\delta_1\uparrow$ & $\delta_2\uparrow$ & $\delta_3\uparrow$ \\ 
\hline \hline
(1) PSMNet \cite{chang2018pyramid}, sample on disparity, $D=64$ &   0.0433   &  0.0252     &  0.2541    &   0.1340       &      0.9722         &     0.9833            &     0.9900             \\

(2) 360SD-Net \cite{wang2020360sd}, sample on disparity, $D=64$ &  0.0387    &   0.0198    &  0.2286    &     0.0955     &     0.9776          &        0.9900       &    0.9940             \\

\hline
(3) PanoDepth (ours), one-level, $D=32$ &   0.0253  &   0.0222 &  0.2268    &   0.0686  & 0.9756 & 0.9874 & 0.9976 \\
(4) PanoDepth (ours), one-level, $D=64$ &   0.0229 &   0.0087 &  0.1731    &   0.0606  & 0.9900 & 0.9969 & 0.9987 \\
\textbf{(5) PanoDepth (ours), two-level, $D_1=48, D_2=24$} &  \textbf{0.0178}    &  \textbf{0.0064}     &  \textbf{0.1415}    &  \textbf{0.0519}   & \textbf{0.9928} & \textbf{0.9976} & \textbf{0.9990} \\
\hline 
\end{tabular}
\end{footnotesize}
\caption{A quantitative comparison between the PanoDepth stereo matching  and existing stereo matching networks on the Omnidirectional stereo dataset \cite{zioulis2019spherical} where up-down stereo pairs are used as input and output the depth of bottom view.  Our proposed stereo matching module (3,4,5) outperforms both (1) PSMNet \cite{chang2018pyramid} and (2) 360SD-Net \cite{wang2020360sd}. The two cascade level setting achieves the best performance.}
\label{tab:ablation_stereo}
\end{center}
\end{table*}

\begin{table*}[hbt!]
\begin{center}
\begin{footnotesize}
\begin{tabular}{m{0.8em} m{4em} m{15em} m{3em} m{3.8em} m{3.5em}  m{3em}  m{4.5em}  m{2.5em}  m{2.5em}  m{2.5em} } 
\hline
  & 1st Stage & 2nd Stage w/ 1 synthesize view & $\#$params & Abs Rel$\downarrow$ & Sq Rel$\downarrow$ & RMSE$\downarrow$ & RMSElog$\downarrow$ & $\delta_1\uparrow$ & $\delta_2\uparrow$ & $\delta_3\uparrow$ \\ 
\hline \hline
(1) & CoordNet & N/A & 6.1M &  0.1264 & 0.0888 & 0.4456 & 0.2084 & 0.8533 & 0.9588 & 0.9813  \\
(2) & RectNet & N/A & 10.8M & 0.1409  & 0.0859 & 0.4568 & 0.2124 & 0.8326 & 0.9518 & 0.9822  \\
(3) & CoordNet & PSMNet \cite{chang2018pyramid}, D=32 & 13.0M & 0.1206 &   0.0833 & 0.4293 & 0.2150 & 0.8671 & 0.9548 & 0.9790  \\
(4) & CoordNet & w/ SWL, one-level, D=32 & 13.0M &  0.1132 & 0.0686 & 0.4077 & 0.1869 & 0.8757 & 0.9652 & 0.9863  \\
(5) & RectNet & w/ SWL, one-level, D=32 & 17.6M &  0.1192  &  0.0775 & 0.4202 & 0.1960 & 0.8655 & 0.9607 & 0.9846 \\
(6) & CoordNet & w/ SWL, two-level, $D_1$=32, $D_2$=16 & 16.6M  & \textbf{0.1040} & \textbf{0.0645} & \textbf{0.3918} & \textbf{0.1827} & \textbf{0.8865} & \textbf{0.9676} & \textbf{0.9875}  \\
(7) & RectNet & w/ SWL, two-level, $D_1$=32, $D_2$=16 & 21.3M  & 0.1138 & 0.0761 & 0.4274 & 0.1961 & 0.8711 & 0.9577  &  0.9837 \\
\hline
\end{tabular}
\end{footnotesize}
\caption{An ablation study of the impact of various combinations of coarse estimation network and stereo matching network on the final performance. The experiments are trained on Stanford2D3D \cite{armeni2017joint}. We use two types of coarse estimation networks, (1) CoordNet, and (2) RectNet
\cite{zioulis2018omnidepth}. We can see that even with one synthesize view, our proposed two-stage \textit{PanoDepth} pipeline (3,4,5,6,7) is able to outperform the one-stage-only methods (1,2). Adding Spherical Warping Layer (SWL) (4,5,6,7) and two cascade levels (6,7) further improves the performance. The experimental results indicate that our two-stage pipeline is model-agnostic under various network settings.
}
\label{tab:ablation_stereo_full}
\end{center}
\end{table*}

\begin{figure*}
    \centering
    \includegraphics[width=17cm]{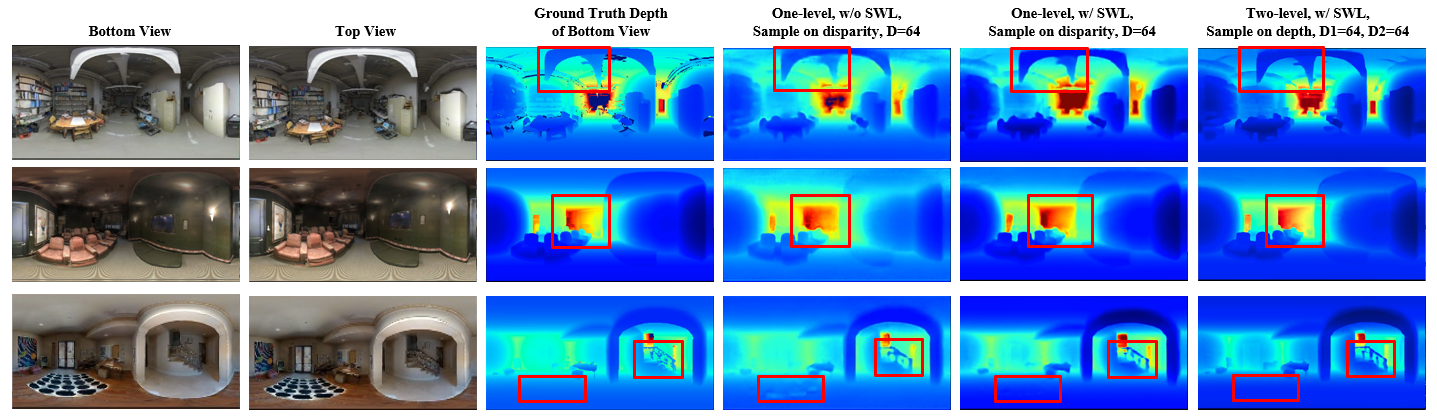}
    \caption{A qualitative comparison to show the effectiveness of SWL in the PanoDepth stereo matching module. We compare between one-level stereo matching method without SWL(4th column), one-level with SWL (5th column), and two-level with SWL (last column). The experiments are trained on Omnidirectional Stereo Dataset \cite{zioulis2019spherical}. Our stereo matching module with SWL recovers clear details and 
   shows fewer artifacts than the one without SWL (see highlighted areas).}
    \label{fig:stereo_compare}
\end{figure*}

\begin{comment}
\begin{figure*}[t!]
\centering
\includegraphics[width=15cm]{figures/real_world_result_v3.PNG}
\caption{A side-by-side qualitative comparison with Bifuse (with fusion) \cite{wang2020bifuse} on three real world images from Laval HDR dataset  \cite{gardner2017illumination}. \textit{PanoDepth} is able to more accurately recover the structure of the room, and capture the details of the surface and the sharp edge more precisely (highlighted with red boxes). We use the pretrained model provided by \cite{wang2020bifuse} to evaluate this dataset.}
\label{fig:real_world}
\end{figure*}
\end{comment}

\textbf{Spherical Warping Layer} 
In order to evaluate the performance of the  \textit{PanoDepth} stereo matching module with the novel Spherical Warping Layer (SWL), we compare it to the state-of-the-art stereo matching approaches, PSMNet \cite{chang2018pyramid} and 360-SD Net \cite{wang2020360sd}, with ground truth up-down 360 stereo pair as input on the Omnidirectional Stereo Dataset \cite{zioulis2019spherical}. 
%We use the official released code, set the number of hypothesis plane to 64, and convert the output disparity into depth. Moreover, to show the effectiveness of our SWL, we run an experiment that do not include SWL or cascade levels (Table \ref{tab:ablation_stereo} (1)). 
In the experiment, we use the officially released code of both approaches, and convert the output disparity into depth for evaluation. We set the number of depth hypothesis planes to 64 to ensure fair and consistent comparison.
We can see from Table \ref{tab:ablation_stereo} that our proposed stereo matching method with SWL outperforms PSMNet \cite{chang2018pyramid} and 360-SD Net \cite{wang2020360sd}, even with one-level setting. Moreover, by adding SWL, our one-level setting outperforms the one without SWL (identical to PSMNet\cite{chang2018pyramid}) by $47\%$ in terms of Abs Rel. with the same 64 sampling planes.
%The experiment with SWL is able to outperform the one without SWL by $47\%$ in terms of Abs Rel. when number of sampling planes are 64. 
%Qualitative comparison of different stereo matching configurations is shown in Figure \ref{fig:stereo_compare}.
Qualitative illustrations of the effectiveness of SWL is shown in Figure \ref{fig:stereo_compare}.
%To study the influence of number of cascade levels as well as hypothesis plane configurations, we conduct extensive experiments (Table \ref{tab:ablation_stereo}). 
%Without the refining level, the one-level setting has an obvious performance drop even with a large number of hypothesis plane ($D_1=64$). At two-level setting, increasing the number of planes is likely to improve the accuracy, but also lead to high GPU memory consumption. We observe that at two-level setting with $D_1=64, D_2=32$, performance slightly drops, which could be a sign of overfitting. Adding more levels could also lead to a similar performance decrease.
%Therefore, we choose a two-level setting with $D_1=48, D_2=24$ as the default since it gives the best performance overall. %In terms of the number of views, we choose two synthesized views as default for our full pipeline, the related ablation study can be found in the Supplemental.

\textbf{Model-agnostic Evaluations}
In Table \ref{tab:ablation_stereo_full}, we further test the performance of the full \textit{PanoDepth} pipeline given different variations of stereo matching networks on Stanford2D3D dataset \cite{armeni2017joint}.
%various combination of networks in both view synthesis and stereo matching stages.
Comparing with the single-stage coarse depth estimation, all two-stage configurations show better performances. By adding a light-weight one-level stereo matching network with 32 depth plane in the second stage, \textit{PanoDepth} can already reach comparable performance to BiFuse \cite{wang2020bifuse}. The performance can be further improved by introducing SWL, adding more cascade levels, and using more sophisticated coarse depth estimation.

In addition, by comparing the performance of two-stage approaches with two different backbones, CoordNet and RectNet, we can also postulate that coarse depth, which has an impact on the synthetic image quality, is positively correlated with the final depth prediction.

More ablation studies regarding the number of synthesized views, the number of hypothesis depth planes, and the comparison between one-stage alternatives (e.g., multi-tasking and adding depth refinement) and our two-stage method can be found in the Appendix.

\begin{comment}
\subsection{Generalization on Real World Data}
\label{subsec:generalization}
To investigate the generalization capability of our method, we evaluate on a 360 dataset that does not have ground truth depth, the Laval indoor HDR database \cite{gardner2017illumination}. The model is pre-trained on 360D \cite{zioulis2018omnidepth} before evaluation. Figure \ref{fig:real_world} shows a side-by-side comparison of our  approach with the current leading approach Bifuse \cite{wang2020bifuse} using its pre-trained model. We can see that our \textit{PanoDepth} generally recovers sharp corners of the rooms much better. %We argue that it might because of the additional geometric constraints brought in by the two-stage pipeline.
\end{comment}

\section{Conclusion and Future Work}
In this paper, we demonstrate a technique that leverages view synthesis and stereo constraints to advance monocular depth estimation performance that can be applied on 360 images.  
We propose a novel model agnostic two-stage framework \textit{PanoDepth} for generating dense high-quality depth from a monocular 360 input.
Extensive experiments show that \textit{PanoDepth} outperforms state-of-the-art approaches by a large margin. Our stereo matching sub-network in the later stage adapts to the 360 geometry and achieves top-ranking performance in 360 stereo matching. We believe the good
performance of PanoDepth could draw more interests from
both the industry and academia to 360 images for its
still under-explored capability in tasks such as depth estimation. We hope our work can
motivate more research and applications in 360 images.
\begin{comment}
\yuyan{Besides these technical innovations, we believe the good
performance of PanoDepth could draw more interests from
both the industry and academia to 360 images for its
still under-explored capability in tasks such as depth estimation. We hope our work can
motivate more research and applications in 360 images. To further facilitate the research in 360 domain, we
plan to make our software and dataset publicly available after the paper is published.}
\end{comment}
There are several research venues we would like to further explore in the future, such as alternative view synthesis methods like \cite{flynn2019deepview, wiles2020synsin}, and 360 depth estimation in outdoor scenarios for applications like autonomous driving.  

%. First, in the synthesis stage, we currently  use a coarse depth estimation network and a DIBR layer to render novel views. We will explore other alternative approaches \cite{flynn2019deepview, wiles2020synsin} that aim to directly synthesize novel views without the intermediate depth map. %Image processing techniques such as depth sharpening \cite{watson2020learning} can be implemented to improve the quality of synthesized views. 
%Second, our framework, as most of other existing approaches, is currently trained and evaluated on indoor 360 datasets. Exploring and extending the use of \textit{PanoDepth} for outdoor scenarios could be very valuable for applications such as autonomous driving.

\section*{Acknowledgments}
\vspace{-0.4cm}
The research of Yuyan Li and Ye Duan were partially supported by the National Science Foundation under award CNS-2018850, National Institute of Health under awards NIBIB-R03-EB028427 and NIBIB-R01-EB02943, and U.S. Army Research Laboratory W911NF2120275. Any opinions, findings, and conclusions or recommendations expressed in this publication are those of the authors and do not necessarily reflect the views of the U.\,S.\ Government or agency thereof.

{\small
\bibliographystyle{ieee_fullname}
\bibliography{egbib}
}

\clearpage

\thispagestyle{empty}
\appendix

\section{Spherical Geometry Analysis}
\label{sec:sph_geometry}
\begin{figure*}[hbt!]
  \begin{center}
    \includegraphics[width=10cm]{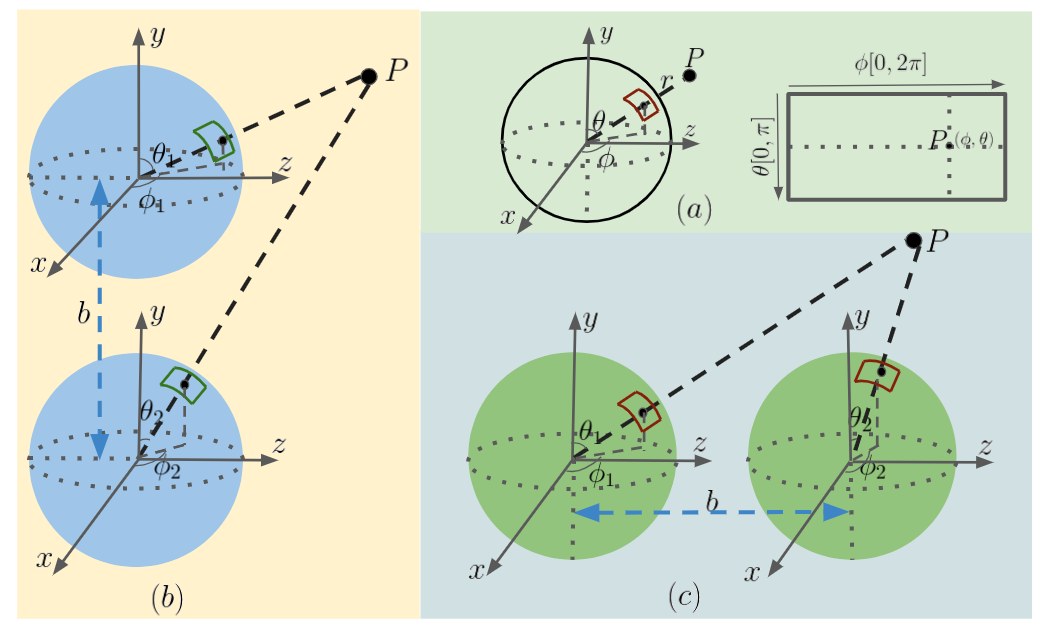}
  \end{center}
  \caption{(a) Projection of a 3D point $P$ from spherical space to equirectangular domain. (b) Vertical spherical disparity model with two cameras placed at baseline $b$. (c) Horizontal spherical disparity model with two cameras placed at baseline $b$.}
  \label{fig:disparity}
\end{figure*}

The projection of a 3D point $P$ from spherical coordinates$(r, \phi, \theta)$ to the ERP image is shown in Figure \ref{fig:disparity}(a), in which $\phi\in [0, \pi]$ represents longitude, and $\theta\in [0, 2\pi]$ is the complement of latitude. $r$ is the distance from $P$ to the camera center, which is considered as depth in our study.
%Spherical stereo model is considered as a set of two cameras positioned at a fixed baseline $b$ in either horizontal or vertical displacement.
Therefore, given two cameras positioned at a fixed baseline $b$, we can build the spherical stereo model. The displacement between the two cameras can be either vertical (Figure \ref{fig:disparity}(b)) or horizontal (Figure \ref{fig:disparity}(c)).

The conversion from a 3D point $P=(x,y,z)$ in Cartesian coordinates to spherical coordinates $(r, \phi, \theta)$ can be calculated as follows:
\begin{equation}
\begin{aligned}
    r &= \sqrt{x^2 + y^2 + z^2}\\[1pt]
    \phi &= arctan(\frac{x}{z})\\[1pt]
    \theta &= arccos(\frac{y}{r})
\end{aligned} 
\label{eq:cart2sph}
\end{equation}
The conversion from spherical to Cartesian is defined as:
\begin{equation}
\begin{aligned}
x &= r\;sin\phi \;sin\theta\\[1pt]
y &= r\;cos\theta\\[1pt]
z &= r\;cos\phi\; sin\theta
\end{aligned} 
\label{eq:sph2cart}
\end{equation}

If we derive the partial derivatives of the conversion process from Equ (\ref{eq:sph2cart}), we would have:
\begin{equation}
\begin{aligned}
    \partial \phi &= \frac{cos\phi}{r\; sin\theta}\partial x -\frac{sin\phi}{r\; sin\theta}\partial z \\
    \partial \theta &= \frac{sin\phi\; cos\theta}{r}\partial x -\frac{sin\theta}{r}\partial y + \frac{cos\phi\; cos\theta}{r}\partial z
\end{aligned}
\label{eq:derivative}
\end{equation}
From Equ (\ref{eq:derivative}), we can calculate the relationship between baseline placement $b=(dx,dy,dz)$ and spherical disparity $d=(d\phi, d\theta)$, for vertical model where $b_y=(0, dy, 0)$:
\begin{equation}
    d \phi = 0, \;\; 
    d \theta = -\frac{sin\theta}{r}b_y
    \label{eq:vertical}
\end{equation}
For horizontal model where $b_x=(dx, 0, 0)$:
\begin{equation}
    d \phi = \frac{cos\phi}{r\; sin\theta}b_x, \;\; 
    d \theta = \frac{sin\phi\; cos\theta}{r}b_x
    \label{eq:horizontal}
\end{equation}
We can see from Equ (\ref{eq:vertical}) that there is no longitudinal displacements for vertical stereo model, while there are both longitudinal and latitudinal displacements for horizontal stereo model as described in Equ (\ref{eq:horizontal}). 
%Therefore, learning vertical disparity is more straightforward than horizontal disparity for a spherical stereo matching network. 
Thus estimating disparity for horizontal stereo model could be difficult with displacements along both horizontal and vertical directions. Our Spherical Warping Layer (SWL) solves this problem by directly working on depth and calculating coordinate mapping based on depth.
In Section \ref{sec:stereo_exp}, we show the quantitative comparison between horizontal and vertical stereo model. 
Based on both the theoretical analysis and experimental results, we choose the vertical stereo as our default setting for the second stage network.

\begin{comment}
\begin{figure*}
    \centering
    \includegraphics[width=13cm]{figures/synthesize_direction.png}
    \caption{Examples of view synthesis along vertical and horizontal directions via Depth-Image-Based Rendering (DIBR).}
    \label{fig:my_label}
\end{figure*}
\end{comment}

%\section{Multi-View Spherical Stereo Matching Network}

\begin{comment}
\section{Network Architecture Detail}

In Table \ref{tab:network_architect_coarse_depth} we show the detail of our coarse depth estimation network. We use CoordNet \cite{zioulis2019spherical} as the baseline architecture, and add an atrous spatial spatial pooling module \cite{chen2017rethinking} to aggregate multi-scale context information. The coarse depth prediction from CoordNet helps to render the original input into novel views via a differential Depth-Image-Based Rendering (DIBR) module. 
The DIBR operation is implemented based on \cite{tulsiani2018layer}. The pixels from source image are splatted on an empty target image, then occlusions are handled by soft z-buffering. The final prediction is the weighted average of points which splat to the same pixel. 

 Table \ref{tab:network_architect_stereo} shows the details of the (cascaded) multi-view spherical stereo matching networt architecture. The depth output from level 1 is used to re-sample depth hypothesis planes in level 2. Based on the depth hypothesis, extracted features are warped and then fused together into one variance cost volume. 
\end{comment}

\begin{figure*}[hbt!]
    \centering
    \includegraphics[width=12cm]{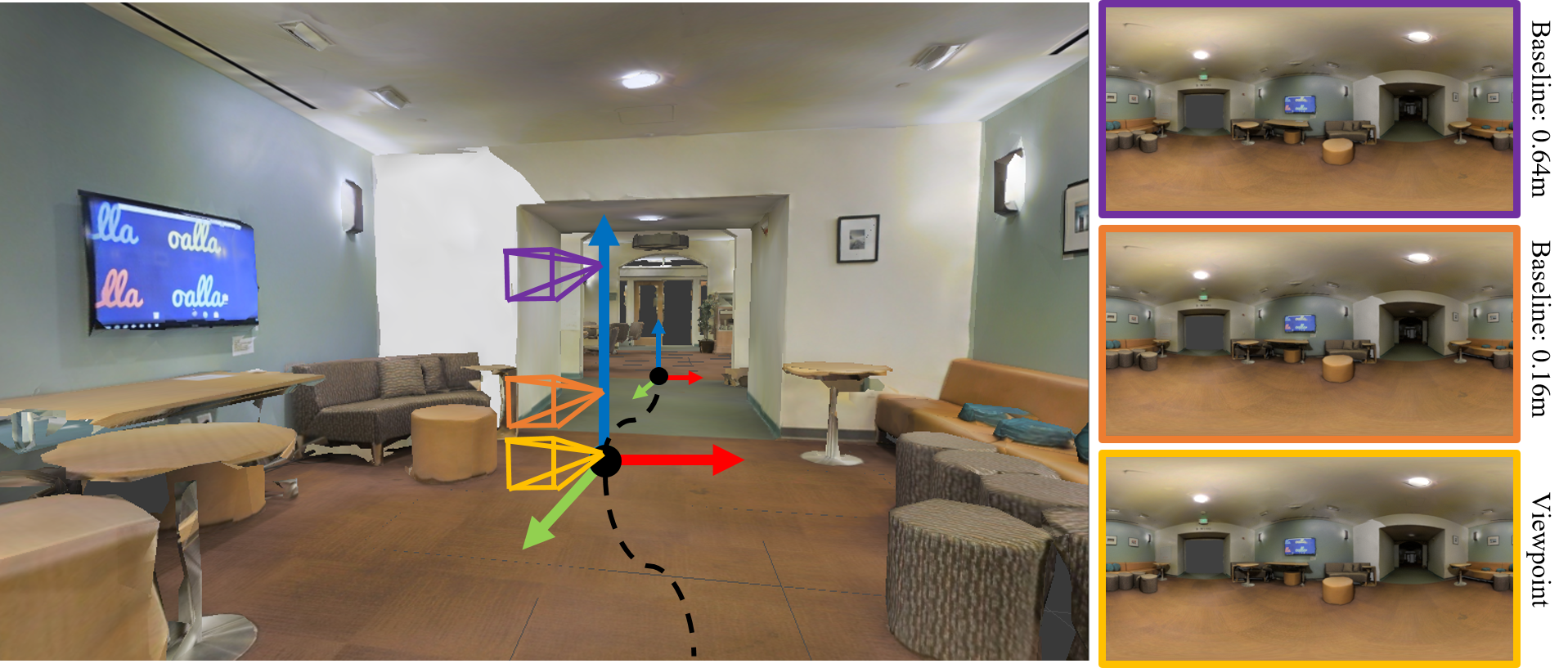}
    \caption{Illustration of scene rendering procedure. Left is the illustration of an example viewpoint (black dot) with three cameras placed along the same vertical axis. On the right side, the three ERP images with purple, orange, and yellow borders correspond to the three camera views (shown in corresponding colors) in the left image. }
    \label{fig:scene_render}
\end{figure*}
\begin{figure*}[hbt!]
    \centering
    \includegraphics[width=12cm]{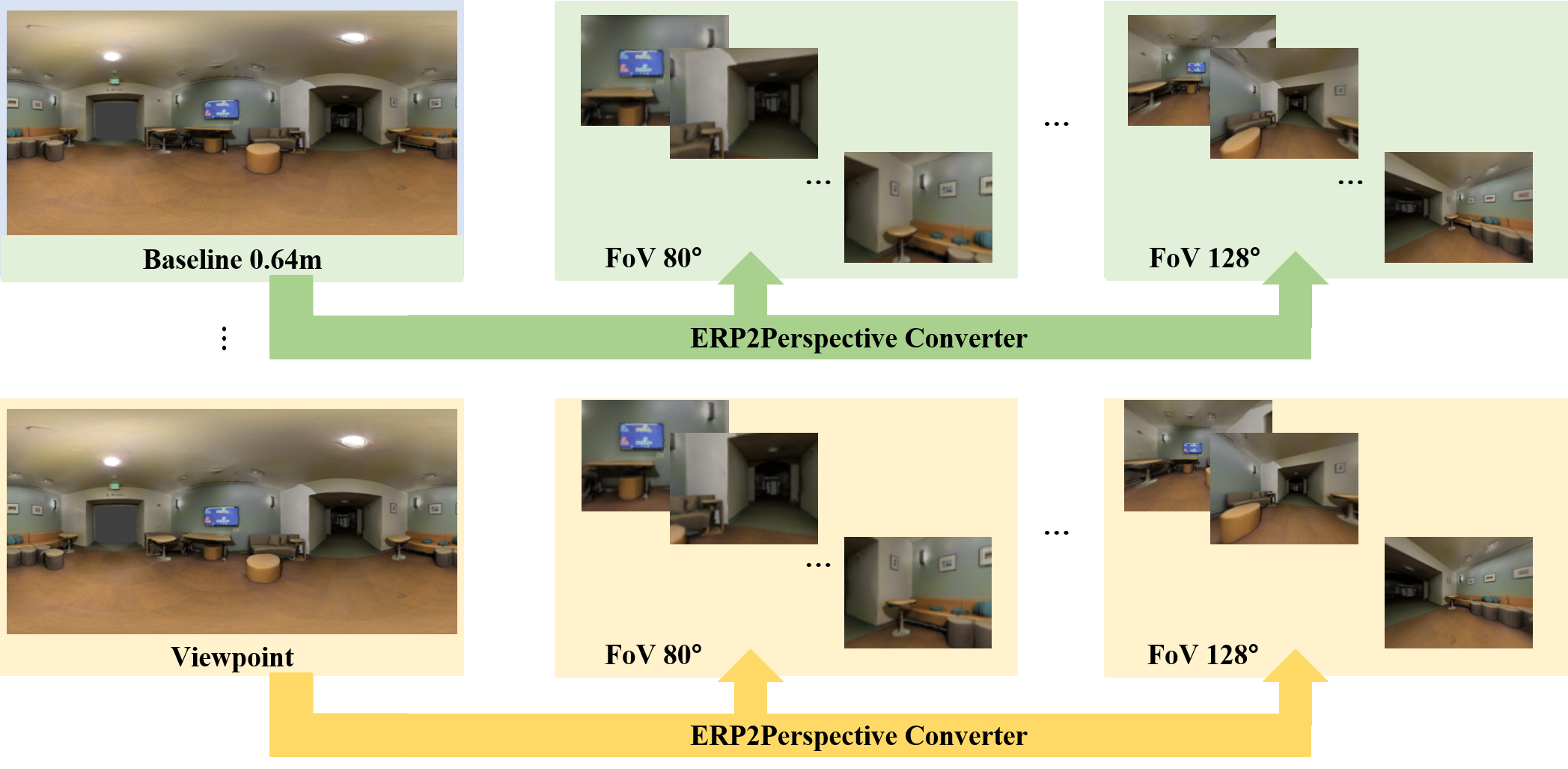}
    \caption{Illustration of the procedure for generating perspective images. For each ERP image rendered with various baselines, we take multiple perspective views using defined field of view (FoV), and map pixels within the FoV to perspective images. The conversion process runs five times for the five FoV settings. }
    \label{fig:pers_gen}
\end{figure*}

\begin{figure*}[hbt!]
    \centering
    \includegraphics[width=16cm]{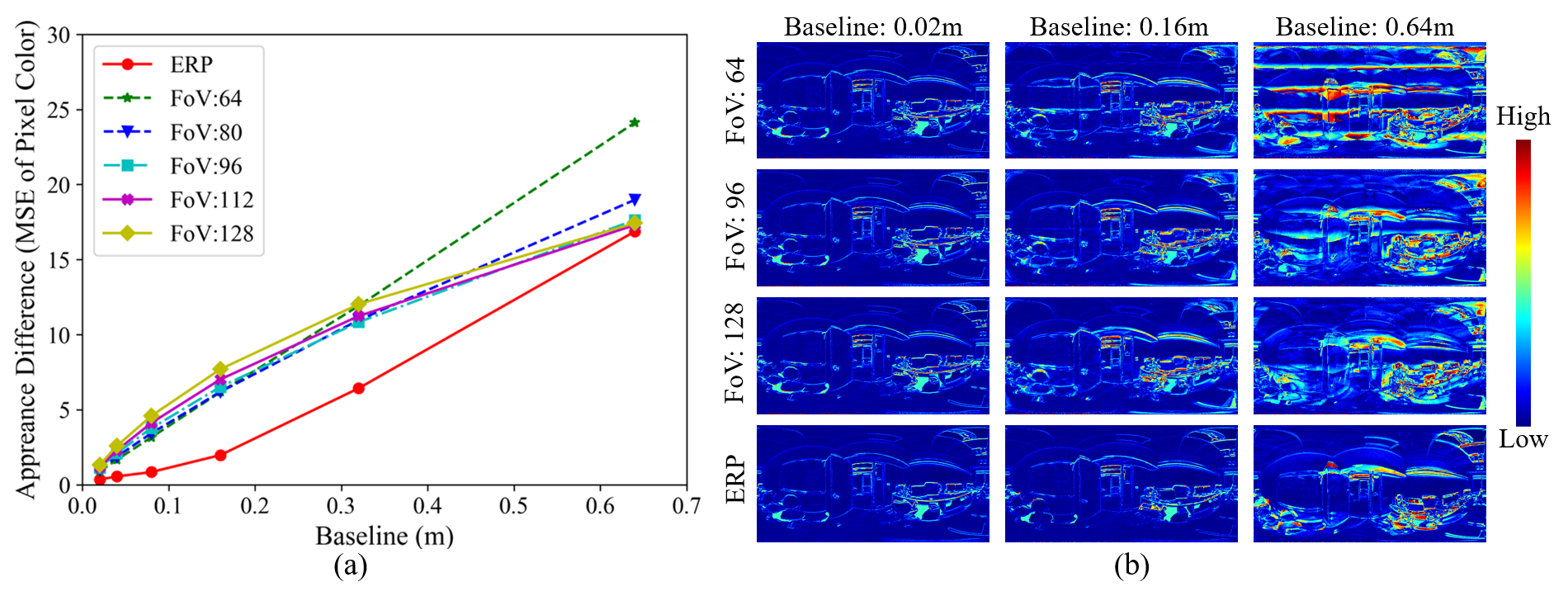}
    \caption{The study of correlations between image FoV, baseline and synthesized view  quality. (a) Plots of MSE error over the baseline settings. (b) Colored error map visualization of MSE for perspective images (row 1,2,3) and ERP images (row 4). }
    \label{fig:baseline_error}
\end{figure*}

\section{Network Architecture and Implementation Details}
In this section, we discuss the implementation details of \textit{PanoDepth} network architecture. \textit{PanoDepth} is model-agnostic, in other words, the sub-networks are replaceable. We only show examples of network implementation with good performance based on our experience.
\subsection{Coarse Depth Estimation Network in View Synthesis Stage}
\begin{figure*}[hbt!]
    \centering
    \includegraphics[width=16cm]{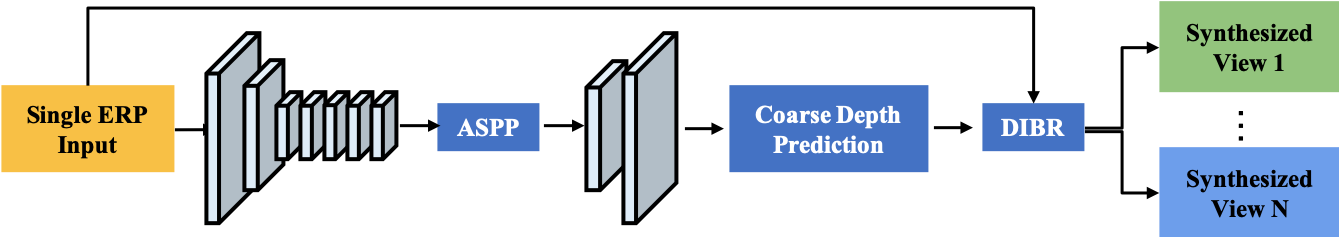}
    \caption{Illustration of view synthesis stage including coarse depth estimation network and DBIR module. }
    \label{fig:view_synthesis_process}
\end{figure*}

For the coarse depth estimation in view synthesis stage, We customize CoordNet \cite{zioulis2019spherical} with ASPP module \cite{chen2017rethinking} and deeper convolution layers. From Table 3 in the main paper, we can see that our customized CoordNet have less parameters than RectNet \cite{zioulis2018omnidepth}, it outperforms RectNet in both one-stage and full two-stage pipeline omnidirectional depth estimation. The detailed network implementation is shown in Table \ref{tab:network_architect_coarse_depth}, the architecture of the CoordNet as well as view synthesis is shown in Figure \ref{fig:view_synthesis_process}. The output coarse depth is then fed into depth image based rendering (DIBR) module together with original RGB 360 image for novel view synthesis given baseline. The DIBR implementation is based on \cite{tulsiani2018layer}. The pixels from source image are splatted on an empty target image, then occlusions are handled by soft z-buffering. The final prediction is the weighted average of points which splat to the same pixel.

Although empirically the quality of view synthesis depends on the accuracy of depth maps, the correlation between the two is not necessarily linear (\cite{luo2018single, wang2020bifuse}). The key for our \textit{model-agnostic} approach is to find the configuration which achieves best performance with reasonable computation resources. We have shown our results with different coarse depth estimation choices as guidelines in main paper Table 3 to justify our choice of network implementation here.

\begin{table*}[hbt!]
\begin{center}
\begin{footnotesize}
\begin{tabular}{m{6em} m{4em} m{4em} m{8em} m{6em} m{6em} m{6em} m{8em}} 
\hline
Layer & Add Coor. & UpSample & Conv. & Batch Norm. & ReLU & Output Layer & Output Dim. \\
\hline
conv1 & Y & - & $7 \times 7, s1, p3, d1$ & Y & Y & conv2,3 & $256 \times 512 \times 16$ \\
conv2 & Y & - & $3 \times 3, s1, p1, d1$ & Y & Y & conv4 & $256 \times 512 \times 16$ \\
conv3 & Y & - & $3 \times 3, s1, p1, d1$ & Y & N & conv4 & $256 \times 512 \times 16$ \\
conv4 & Y & - & $3 \times 3, s2, p1, d1$ & Y & Y & conv5,6 & $128 \times 256 \times 32$ \\
conv5 & Y & - & $3 \times 3, s1, p1, d1$ & Y & Y & conv7,8 & $128 \times 256 \times 32$ \\
conv6 & Y & - & $3 \times 3, s1, p1, d1$ & Y & N & conv7,8 & $128 \times 256 \times 32$ \\
conv7 & Y & - & $3 \times 3, s1, p1, d1$ & Y & Y & conv9,10 & $128 \times 256 \times 32$ \\
conv8 & Y & - & $3 \times 3, s1, p1, d1$ & Y & N & conv9,10 & $128 \times 256 \times 32$ \\
conv9-10 & - & - & repeat conv7-8 & - & - & - & $128 \times 256 \times 32$ \\
conv11 & Y & - & $3 \times 3, s2, p1, d1$ & Y & Y & conv12,13 & $64 \times 128 \times 64$ \\
conv12 & Y & - & $3 \times 3, s1, p1, d1$ & Y & Y & conv14,15 & $64 \times 128 \times 64$ \\
conv13 & Y & - & $3 \times 3, s1, p1, d1$ & Y & N & conv14,15 & $64 \times 128 \times 64$ \\
conv14 & Y & - & $3 \times 3, s1, p1, d1$ & Y & Y & conv16,17 & $64 \times 128 \times 64$ \\
conv15 & Y & - & $3 \times 3, s1, p1, d1$ & Y & N & conv16,17 & $64 \times 128 \times 64$ \\
conv16-37 & - & - & repeat conv14-15 & - & - & - & $64 \times 128 \times 64$ \\
ASPP conv38 & - & - & $1 \times 1, s1, p1, d1$ & Y & - & concat44 & $64 \times 128 \times 256$ \\
ASPP conv39 & - & - & $3 \times 3, s1, p6, d6$ & Y & - & concat44 & $64 \times 128 \times 256$ \\
ASPP conv40 & - & - & $3 \times 3, s1, p12, d12$ & Y & - & concat44 & $64 \times 128 \times 256$ \\
ASPP conv41 & - & - & $3 \times 3, s1, p18, d18$ & Y & - & concat44 & $64 \times 128 \times 256$ \\
avgpool42 & - & - & - & - & - & conv43 & - \\
ASPP conv43 & - & - & $1 \times 1, s1, p1, d1$ & Y & - & concat44 & $1 \times 1 \times 256$ \\
concat44 & - & - & - & - & - & conv45 & - \\
ASPP conv45 & - & - & $1 \times 1, s1, p1, d1$ & Y & - & conv46 & $64 \times 128 \times 64$ \\
conv46 & - & 2.0 & $3 \times 3, s1, p1, d1$ & Y & Y & conv47 & $128 \times 256 \times 32$ \\
conv47 & - & - & $3 \times 3, s1, p1, d1$ & Y & Y & conv48 & $128 \times 256 \times 32$ \\
conv48 & - & 2.0 & $3 \times 3, s1, p1, d1$ & Y & Y & conv49 & $256 \times 512 \times 16$ \\
conv49 & - & - & $3 \times 3, s1, p1, d1$ & Y & Y & conv50 & $256 \times 512 \times 16$ \\
conv50 & Y & - & $7 \times 7, s1, p3, d1$ & N & N & - & $256 \times 512 \times 1$ \\
\hline
\end{tabular}
\end{footnotesize}
\end{center}
\caption{Our customized CoordNet has 50 convolution layers with ASPP module. Each properties (s, p, d) means (stride, padding, dilation) in the convolutional block.}
\label{tab:network_architect_coarse_depth}
\end{table*}

\subsection{Stereo Matching Network}
\begin{figure*}[hbt!]
    \centering
    \includegraphics[width=16cm]{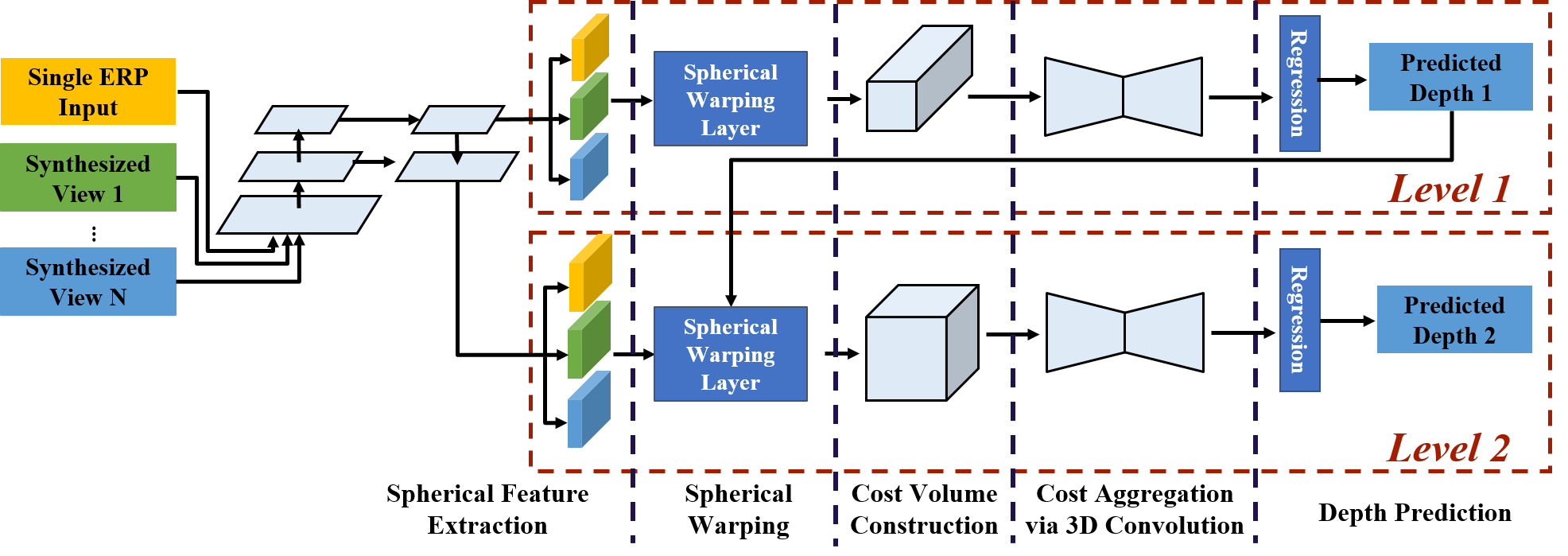}
    \caption{Illustration of stereo matching network architecture with feature extraction, spherical warping, cost volume construction, cost aggregation and depth prediction. }
    \label{fig:stereo_matching_process}
\end{figure*}

Our stereo matching network adapts from PSMNet, with Spherical Warping Layer (SWL) and other features enabled by SWL such as multi-view cost fusion and cascade design. The architecture of our stereo matching network is illustrated in Figure \ref{fig:stereo_matching_process}. Note that as \textit{PanoDepth} is model-agnostic, we can replace PSMNet with other deep stereo matching networks simply by integrating SWL in cost volume construction.

During the construction of the cost volume, it is preferable that the inverse depth of the  depth hypothesis planes are uniformly distributed. Otherwise, the estimated depth from the spherical stereo matching network could be biased. To ensure the uniform distribution of depth hypothesis planes, we develop the following process. The first step is to generate depth hypothesis planes. As mentioned in our main paper, cascade mechanism is incorporated for more accurate stereo matching.
%Each cascade level has different number of depth hypothesis planes.
%There are multiple levels $(l \in {0,1,...N})$ of depth hypothesis planes due to the proposed cascaded mechanism. 
At the first cascade level $l=1$, we sample uniformly on the inverse depth domain:
\begin{equation}
    \frac{1}{d_j} = \frac{1}{d_{max}} + (\frac{1}{d_{min}}-\frac{1}{d_{max}})\frac{v_0\times j}{D-1}, j=0,1,..., D-1
    \label{eq:hypothesis}
\end{equation}
where $D$ is the total number of hypothesis planes at level $l=1$, $d_j$ is the $j^{th}$ depth plane, $d_{min}$ and $d_{max}$ are the minimum and maximum of the depth image, $v_0$ is the initial plane interval.
At a subsequent level $l(l>1)$, hypothesis plane ranges are adjusted according to the prediction from the previous level. With an initial interval $v_l$, we update the minimum and maximum of depth and re-calculate the interval for each individual pixel:
\begin{equation}
    \begin{aligned}
        d'_{min} &= max(\frac{1}{\tilde{d}_{l-1}} - \frac{D_l\times v_l}{2},\;\;\frac{1}{d_{max}}) \\
        d'_{max} &= min(\frac{1}{d_{min}},\; \frac{1}{\tilde{d}_{l-1}} + \frac{D_l\times v_l}{2})\\
        v'_l &= (\frac{1}{d'_{min}} - \frac{1}{d'_{max}})/(D_l-1)
    \end{aligned}
\end{equation}
where $D_l$ is the number of hypothesis planes for the current level, $\tilde{d}_{l-1}$ is the depth prediction from level $l-1$, and $v'_l, d'_{min}, d'_{max}$ is the updated plane interval, minimum and maximum depth  respectively. Then the depth plane hypothesis $d_{j, l}$ for level $l$ can be calculated based on Equ (\ref{eq:hypothesis}). 

In the next step, we warp the feature map of each reference image to the target image at each hypothesis plane. This warping is calculated using coordinate mapping between the reference and target image, determined by spherical disparity:

%To tackle this, 360-SD Net\cite{wang2020360sd} proposed a learnable cost volume (LCV) with a 1D convolution filter, which searches the optimal disparity interval in order to construct the optimal cost volume. However, based on our observation, the vertical disparity can be generated in a deterministic way given depth hypothesis and latitudinal value $\theta$. Therefore, training additional parameters such as LCV for this process would not be necessary. Under this premise, we propose a novel differentiable Spherical Warping Layer, where we transform the depth hypothesis to disparity, i.e., the coordinate mapping between the reference image and the target image. In addition, our spherical warping layer can be directly used to handle cascaded stereo matching network, while there is no trivial adaptation for LCV to fit cascaded scheme. Given the depth hypothesis $d_{j,l}$ from Equation \ref{eq:hypothesis}, the transformation can be determined as:
\begin{equation}
   C_{x}=0, \; C_{y} = \frac{cos(\theta)\times baseline}{d_{j,l}} \times \frac{H}{\pi} 
   \label{eq:sw}
\end{equation}
where $H$ is the height of input feature map. Note that the range of $\theta$ is normalized into $[-\pi/2, \pi/2]$

\section{Experiments}

\subsection{View Synthesis Analysis}
\label{sec:view_analysis}

To analyze the correlation between image synthesis quality, image field of view (FoV), and stereo baselines, a dataset consists of various combinations of image FoV and baselines is required. Since there is no such benchmark dataset available, we construct our own dataset by rendering 3D meshes from a real-world dataset Stanford2D3D \cite{armeni2017joint}. %Specifically, we define a virtual camera as well as a virtual point light that follows a customized path using Blender \cite{blender} to render panoramic images from 3D scenes. 
In our dataset, images are organized as two parts, equirectangular projection (ERP) and perspective. 
For each 3D mesh, we first generate the ERP dataset using Blender \cite{blender} to render panoramic image sequence at virtual camera viewpoints sampled from a customized path. At each viewpoint, we render RGB-D ERP images with cameras placed along the vertical axis with baseline displacements of $b\in\{0m, 0.01m, 0.02m, 0.04m, 0.08m, 0.16m, 0.32m, 0.64m\}$. An illustration of scene rendering is shown in Figure \ref{fig:scene_render}.

We collect results from over 1000 viewpoints, in total more than 8000 RGB-D ERP images including original view point and 7 vertical baseline settings in the ERP dataset. Using these rendered RGB-D ERP images, we then build the perspective dataset given five FoV settings, $\{{64}^\circ, {80}^\circ, {96}^\circ, {112}^\circ, {128}^\circ\}$. The ERP dataset can also be considered as a special case of the FoV setting of ${360}^\circ$. Given a ERP RGB-D image and the FoV settings, we uniformly sample $4\times 6$ pairs of $\theta$ and $\phi$ angles from the spherical space. For each pair of $\theta$ and $\phi$ angles, we can define a homography transformation $H$ from spherical space to perspective image space with the FoV setting. By mapping the pixels from ERP image to perspective image using transformation $H$, and filling the holes by image interpolation, the perspective image is generated.
%We use the library \cite{py360convert} for the conversion from ERP image to perspective images.
In total, around 2 million perspective images are collected. Figure \ref{fig:pers_gen} shows an illustration.
%We collect $1000+$ sets of RGB-D panorama images, with each set contains 7 RGB-D images taken at different vertical baseline displacements. Next, we project ERP images onto perspective images. For each RGB-D ERP image, we convert it to sequences of perspective images. There are five FoV settings, $\{{64}^\circ, {80}^\circ, {96}^\circ, {112}^\circ, {128}^\circ\}$ in our dataset.

Using the constructed ERP and perspective datasets, we analyze the correlation between image FoV, baseline displacement, and view synthesis quality. 
In detail, we synthesize novel views for each viewpoint via Depth-Image-Based Rendering (DIBR) \cite{tulsiani2018layer} for both perspective images and ERP images with their ground truth depth maps. To make a fair comparison, we stitch the synthesized perspective images as ERP format and eliminate the image stitching error during comparison. The error metric used in the comparison is the Mean Squared Error(MSE).

Figure \ref{fig:baseline_error} shows both the qualitative and quantitative results of the experiment. In Figure \ref{fig:baseline_error}, we can see that synthesized ERP images in general have the least errors and artifacts compared with perspective images at different FoV settings. There are two major sources of errors in view synthesis of perspective images: boundary effect and object occlusions. Unlike perspective images, given ${360}^\circ$ horizontal and ${180}^\circ$ vertical FoV in one image, 360 images are free of boundary effect. From the experiment, we can also observe that when the baseline increases, ERP and large FoV images are less effected. We conclude that ERP images have advantages in novel view synthesis, especially with large baselines. Such advantage can furthermore benefit the stereo matching stage for better depth prediction.
%when the baselines are smaller, the dominant source of the view synthesis error is the boundary effect,  
%The object occlusion error increases slightly when baseline grows, but is not affected by changes of FoV. The boundary effect increases with either smaller FoVs or larger baselines. 
%To conclude, with greater FoV the synthesized views are less sensitive to large baselines, and synthesized ERP images have the least error and artifacts.

\subsection{Ablation Studies}

\subsubsection{Convergence Analysis}

\begin{figure*}[hbt!]
    \centering
    \includegraphics[width=16cm]{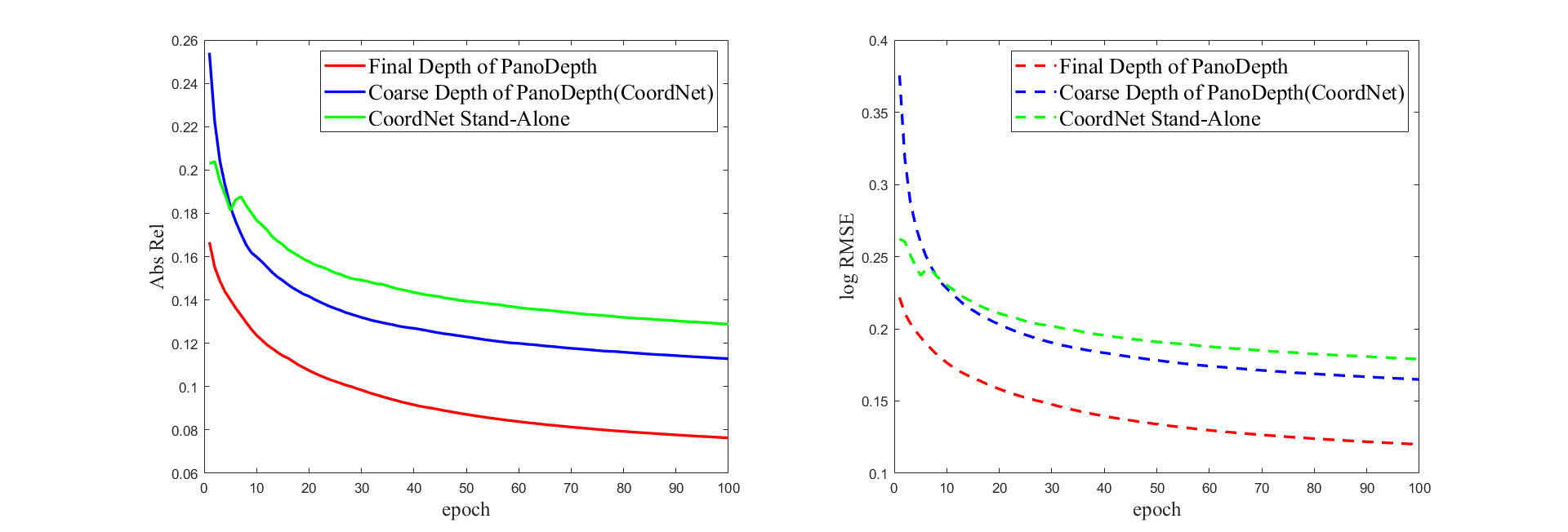}
    \caption{Depth prediction error metrics of the coarse depth estimation(first stage) and the final output(second stage) over epochs in terms of Abs Rel (left) and log RMSE (right). }
    \label{fig:convergence}
\end{figure*}

In Figure \ref{fig:convergence}, we plot the depth error in the form of Abs Rel and RMSE(log) for both the first stage and the final output. To demonstrate the advantage of end-to-end training, we add the results from monocular depth estimation with CoordNet \cite{zioulis2019spherical} as backbone (stand-alone CoordNet) to the plot. The experiment is conducted on Omnidirectional Stereo Dataset \cite{zioulis2019spherical} where the left-down view is used for training and testing. 
We can see that the coarse depth estimation network converges fast at earlier epochs. This enables the second stage stereo matching network to converge. Moreover, since our pipeline is en-to-end, the gradient of the second stage flows back to the view synthesis stage and facilitates the generation of better quality novel synthesized views. We compare a stand-alone CoordNet with the CoordNet that is integrated in our full pipeline. As shown in Figure \ref{fig:convergence}, the stand-alone version has the largest depth prediction error, which indicates that simultaneous training for networks in two stage is beneficial for convergence.
%However, since stereo matching result depends on the quality of synthesized views, low-quality synthesized views from early iterations may cause the gradient explosion in the stereo matching network. During the training of our end-to-end pipeline, we freeze the parameters of the second stage network while only the parameters of the first stage network are optimized for the first 300th iterations in the beginning. 

\subsubsection{Number of Synthesized Views}
In Table \ref{tab:number_of_view}, we show the impact of the number of synthesized views on performance of stereo matching. The experiment is conducted on Omnidirectional Stereo Dataset \cite{zioulis2019spherical} where the left-down view is used for training and testing. 
We use the ground truth depth to synthesize multiple novel views on different baseline settings.
Comparing with one novel view, more stereo constraints are brought in when given two or more synthesized views. We observe that there is a performance drop when using four views. As we explored in our analysis in Section \ref{sec:view_analysis}, when the baseline becomes larger, it introduces more errors in view synthesis. Thus, adding too many synthesized views could accumulate these errors, which could lead to performance drop. Hence in this paper, we choose three synthesized views as the default setting.

\begin{table*}[hbt!]
\begin{center}
\begin{footnotesize}
\begin{tabular}{ m{18em}  m{4em} m{4em}  m{4em}  m{5em}  m{4em}  m{4em}  m{4em} } 
\hline
$\#$ Synthesized Views(baseline(s)) & Abs Rel$\downarrow$ & Sq Rel$\downarrow$ & RMSE$\downarrow$ & RMSElog$\downarrow$ & $\delta_1\uparrow$ & $\delta_2\uparrow$ & $\delta_3\uparrow$ \\ 
\hline
1 View ($0.12m$) &    0.0849 & 0.0373 & 0.3180 & 0.1332 & 0.9267 & 0.9860 & 0.9961
  \\
1 View ($0.24m$) &    0.0845 & 0.0370 & 0.3164 & 0.1323 & 0.9274 & 0.9862 & 0.9963  \\
\hline
2 Views ($-0.12m$, $0.12m$) &   \textbf{0.0829} & 0.0364 & 0.3147 & 0.1311 & 0.9296 & 0.9864 & 0.9962  \\
2 Views ($-0.24m$, $0.24m$) &   0.0842 & 0.0372 & 0.3187 & 0.1328 & 0.9274 & 0.9859 & 0.9961  \\
\hline
3 Views ($-0.12m$,$0.12m$,$0.24m$) &   0.0830 & \textbf{0.0361} & \textbf{0.3129} & \textbf{0.1305} & \textbf{0.9302} & \textbf{0.9867} & \textbf{0.9963} \\
3 Views ($-0.24m$,$0.24m$,$0.4m$) &   0.0839 & 0.0365 & 0.3141 & 0.1316 & 0.9288 & 0.9866 & 0.9962   \\
\hline
4 Views ($-0.24m$,$-0.12m$,$0.12m$,$0.24m$) &   0.0834 & 0.0365 & 0.3152 & 0.1316 & 0.9287 & 0.9864 & 0.9962
\\
4 Views ($-0.4m$,$-0.24m$,$0.24m$,$0.4m$) &   0.0840 & 0.0372 & 0.3175 & 0.1322 & 0.9281 & 0.9862 & 0.9961
 \\
\hline
\end{tabular}
\end{footnotesize}
\caption{An ablation study of the number of synthesized views and the baselines. All the views are generated along the vertical axis with predefined baselines.}
\label{tab:number_of_view}
\end{center}
\end{table*}

\subsubsection{Number of Hypothesis Planes}
Table \ref{tab:number_of_plane} shows the ablation study of stereo matching network using different cascade levels as well as different number of hypothesis planes. Adding cascade level and  planes improves the performance, but also increases the computation cost. The experiment is trained and tested on Omnidirectional Stereo Dataset \cite{zioulis2019spherical} where bottom-up stereo pairs are used.

\begin{table*}[hbt!]
\begin{center}
\begin{footnotesize}
\begin{tabular}{ m{3em} m{12em}  m{4em}  m{4em}  m{4em}  m{6em}  m{3em}  m{3em}  m{3em}} 
\hline
Cascade Level(s) & Number of Hypothesis Planes & Abs Rel$\downarrow$ & Sq Rel$\downarrow$ & RMSE$\downarrow$ & RMSElog$\downarrow$ & $\delta_1\uparrow$ & $\delta_2\uparrow$ & $\delta_3\uparrow$ \\
\hline
1 & $D=32$ & 0.0253 & 0.0222 & 0.2268 & 0.0686 & 0.9756 & 0.9874 & 0.9976 \\
1 & $D=48$ & 0.0236 & 0.0102 & 0.1860 & 0.0626 & 0.9854 & 0.9960 &  0.9984\\
1 & $D=64$ & 0.0229 & 0.0087 & 0.1731 & 0.0606 & 0.9900 & 0.9969 &  0.9987\\
2 & $D_1=32, D_2=16$ & 0.0186 & 0.0069 & 0.1491 & 0.0544 & 0.9920 & 0.9974 & 0.9989 \\
2 & $D_1=48, D_2=24$ & 0.0178 & 0.0064 & 0.1415 & 0.0519 & 0.9928 & 0.9976 &  0.9990\\
2 & $D_1=64, D_2=32$ & \textbf{0.0166} & \textbf{0.0062} & \textbf{0.1419} & \textbf{0.0513} & \textbf{0.9933} & \textbf{0.9976} & \textbf{0.9989} \\
\hline
\end{tabular}
\end{footnotesize}
\end{center}
\caption{An ablation study of different number of cascade levels, and the number of hypothesis planes. We see that two-level outperforms one-level methods, and more hypothesis planes improve the stereo matching performance.}
\label{tab:number_of_plane}
\end{table*}

\subsubsection{One-stage VS. Two-stage}
In this section, we compare the performance between one-stage and two-stage 360 monocular depth estimation approach. Besides the vanilla one-stage approach, we also add several one-stage approaches with techniques that utilizing multi-tasking or additional refinement module to improve depth quality. For multi-tasking, there are methods that combines normal and boundary estimation \cite{eder2019pano}, or semantic segmentation and layout prediction \cite{zeng2020joint}, with depth estimation task. 
%Layout and semantic annotations of 360 images are expensive to require, while normal maps can be directly derived from depth maps. 
%We obtain normal maps from ground truth and use it as additional supervision signal for multi-tasking ablation study. In detail, we add another decoder in  CoordNet that estimates normal map. The loss is calculated as the summation of the supervision loss of both depth and normal maps.
In this experiment, we train a multi-task depth network with a shared weight encoder and two decoders for both depth and normal maps prediction. 
For refinement, we append another CoordNet to the first CoordNet as refinement network, which takes the original RGB and the coarse depth estimation as input and outputs a refined depth map.
%We conducted two ablations on Stanford2D3D \cite{armeni2017joint}, to show that our \textit{PanoDepth} is able to achieve the best performance. We use a vanilla CoordNet as a baseline method. In the first experiment, we append another CoordNet to the first CoordNet as refinement network, which takes the original RGB and the coarse depth estimation as input and outputs a refined depth map. In the second experiment, we add another decoder in the CoordNet that estimates normal map. The loss is calculated as the summation of the supervision loss of both depth and normal maps.
The results is shown in Table \ref{tab:additional_constrain}. Adding a refinement network boosts the performance by 7.8\% (Abs Rel). Multi-tasking improves the performance by 15.9\% (Abs Rel). However, multi-tasking usually requires additional training labels which is not suitable for many cases where only ground truth depth is available. Our \textit{PanoDepth} is able to outperform the vanilla CoordNet by 18.5\%, which has the best performance. Besides, since PanoDepth is model-agnostic, we can even add multi-tasking to the coarse depth estimation network to further improve the performance of the full pipeline. 

\begin{table*}[hbt!]
\begin{center}
\begin{footnotesize}
\begin{tabular}{ m{12em} m{4em} m{4em}  m{4em}  m{4em}  m{6em}  m{3em}  m{3em}  m{3em}} 
\hline
 Methods & \#params &  Abs Rel$\downarrow$ & Sq Rel$\downarrow$ & RMSE$\downarrow$ & RMSElog$\downarrow$ & $\delta_1\uparrow$ & $\delta_2\uparrow$ & $\delta_3\uparrow$ \\
\hline
CoordNet & 6.1M &  0.1264 & 0.0888 & 0.4456 & 0.2084 & 0.8533  & 0.9558 & 0.9813 \\
CoordNet with Refinement & 12.2M &  0.1165 & 0.0794 & 0.4186 & 0.1927 & 0.8668 & 0.9615 & 0.9852 \\
CoordNet with multi-tasking & 7.2M &  0.1063 & 0.0695 & 0.4089 & 0.1877 & \textbf{0.8841} & 0.9677 &  0.9877 \\
PanoDepth &  16.6M & \textbf{0.1030} & \textbf{0.0635} & \textbf{0.4001} & \textbf{0.1828} & 0.8821 & \textbf{0.9682} & \textbf{0.9877} \\

\hline
\end{tabular}
\end{footnotesize}
\end{center}
\caption{Performance comparisons between CoordNet, CoordNet with refinement, CoordNet with depth and normal prediction (multi-tasking), and our {PanoDepth}. {PanoDepth} is able to outperform the vanilla CoordNet by 18.5\%, achieve the best performance comparing to other techniques such as adding refinement or multi-tasking. For PanoDepth stereo network setting, we use two-level cascade ($D_1 = 48, D_2=24$).}
\label{tab:additional_constrain}
\end{table*}

\begin{table*}[hbt!]
\begin{center}
\begin{footnotesize}
\begin{tabular}{ m{16em}  m{4em} m{4em}  m{4em}  m{6em}  m{4em}  m{4em}  m{4em} } 
\hline
Positions of stereo and baseline(s) & Abs Rel$\downarrow$ & Sq Rel$\downarrow$ & RMSE$\downarrow$ & RMSElog$\downarrow$ & $\delta_1\uparrow$ & $\delta_2\uparrow$ & $\delta_3\uparrow$ \\ 
\hline
V $0.24m$, V $-0.24m$ &  \textbf{0.0553} & \textbf{0.0220} & \textbf{0.2561} & \textbf{0.0995} & \textbf{0.9625} & \textbf{0.9926} & \textbf{0.9977} \\

H $0.24m$, H $-0.24m$ & 0.0937 & 0.0415 & 0.3496 & 0.1403 & 0.9057 & 0.9850 & 0.9966  \\

H $0.24m$, V $0.24m$&  0.0881 & 0.0365 & 0.3190 & 0.1339 & 0.9279 & 0.9863 & 0.9961  \\
\hline
\end{tabular}
\end{footnotesize}
\end{center}
\caption{Evaluations of multi-view stereo matching network on different horizontal and vertical spherical stereo settings. V represents vertical displacement, and H represents horizontal displacements. The vertical stereo setting (top row) achieves the best performance.}
\label{tab:stereo_direction}
\end{table*}

\begin{table*}[hbt!]
\begin{center}
\begin{footnotesize}
\begin{tabular}{ m{20em}  m{4em} m{4em}  m{4em}  m{6em}  m{4em}  m{4em}  m{4em} } 
\hline
Sampling method & Abs Rel$\downarrow$ & Sq Rel$\downarrow$ & RMSE$\downarrow$ & RMSElog$\downarrow$ & $\delta_1\uparrow$ & $\delta_2\uparrow$ & $\delta_3\uparrow$ \\ 
\hline 
Random sampling &  0.0818 & 0.0410 & 0.3198 & 0.1204 & 0.9199 & 0.9893 & 0.9972  \\

Uniform depth sampling &  0.0662 & 0.0334 & \textbf{0.2559} & 0.1065 & 0.9625 & 0.9922 & 0.9962 \\

Uniform reverse depth sampling (proposed) & \textbf{0.0553} & \textbf{0.0220} & 0.2561 & \textbf{0.0995} & \textbf{0.9625} & \textbf{0.9926} & \textbf{0.9977}  \\
\hline

\end{tabular}
\end{footnotesize}
\end{center}
\caption{Depth sampling methods comparison. We compare three sampling methods, random sampling, uniform depth sampling, and uniform reverse depth sampling. The uniform reverse depth sampling from our proposed Spherical Warping Layer performs the best.}
\label{tab:sampling}
\end{table*}

\subsubsection{Spherical Stereo Model}
\label{sec:stereo_exp}
To compare the horizontal and vertical stereo model performance, we use the ground truth depth to generate synthesized views via DIBR with both horizontal and vertical baselines as inputs. We then evaluate the multi-view spherical stereo matching results using two synthesized views with different combinations of spherical stereo model, including i) both horizontal, ii) both vertical and iii) one horizontal + one vertical. As listed in Table \ref{tab:stereo_direction}, stereo model with all vertical directions has the best performance. This is consistent with the spherical geometry analysis in Section \ref{sec:sph_geometry}. The experiment is performed on Omnidirectional Stereo Dataset \cite{zioulis2019spherical} where only left-down view is used. The results are all collected at the 60th epoch as a fair comparison. 
%Adding one horizontal view or using both horizontal views will leads to performance drop. 

\subsubsection{Depth Sampling Analysis}

To validate the effectiveness of uniform sampling on reverse depth in our proposed spherical warping layer, we compare it with other sampling strategy in Table \ref{tab:sampling} including random sampling and uniform depth sampling. The numbers of hypothesis planes $D_1$ and $D_2$ are 48 and 24 respectively. For random sampling, we perturb a small Gaussian noise on the uniformly sampled depth hypothesis. From Table \ref{tab:sampling}, we see that our proposed uniform reverse depth sampling has the best performance.  The experiment is performed on Omnidirectional Stereo Dataset \cite{zioulis2019spherical} where only left-down view is used. The results are all collected at the 60th epoch as a fair comparison.

\begin{figure*}
    \centering
    \includegraphics[width=17cm]{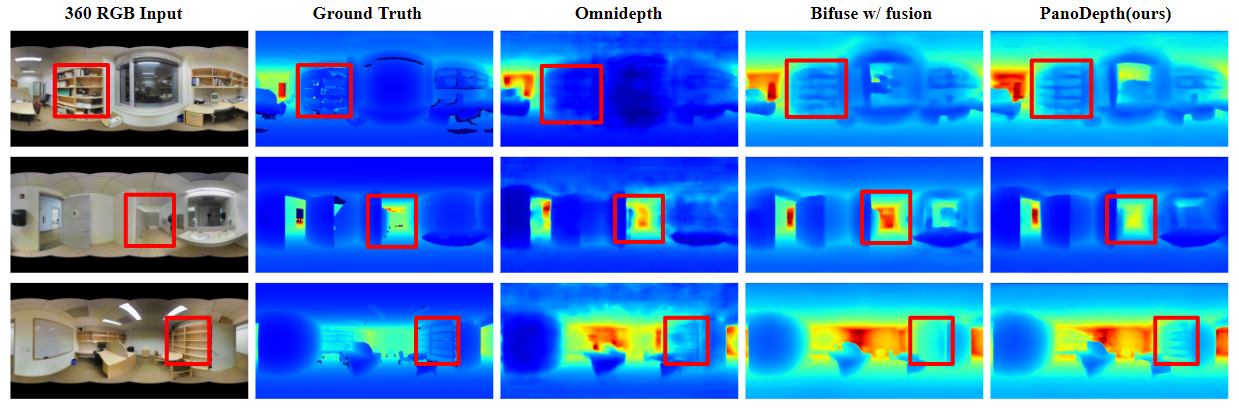}
    \caption{Qualitative comparisons with the state-of-the-art approaches on Stanford2D3D \cite{armeni2017joint}. We show the results of RectNet  proposed in Omnidepth \cite{zioulis2018omnidepth} (3rd column), BiFuse with Fusion \cite{wang2020bifuse} (4th column), and our PanoDepth (5th column). We use red boxes to highlight the differences between predicted depth maps.}
    \label{fig:360d}
\end{figure*}

\subsection{Additional Qualitative Comparisons with State-of-the-art Algorithms}

\begin{figure*}
    \centering
    \includegraphics[width=17cm]{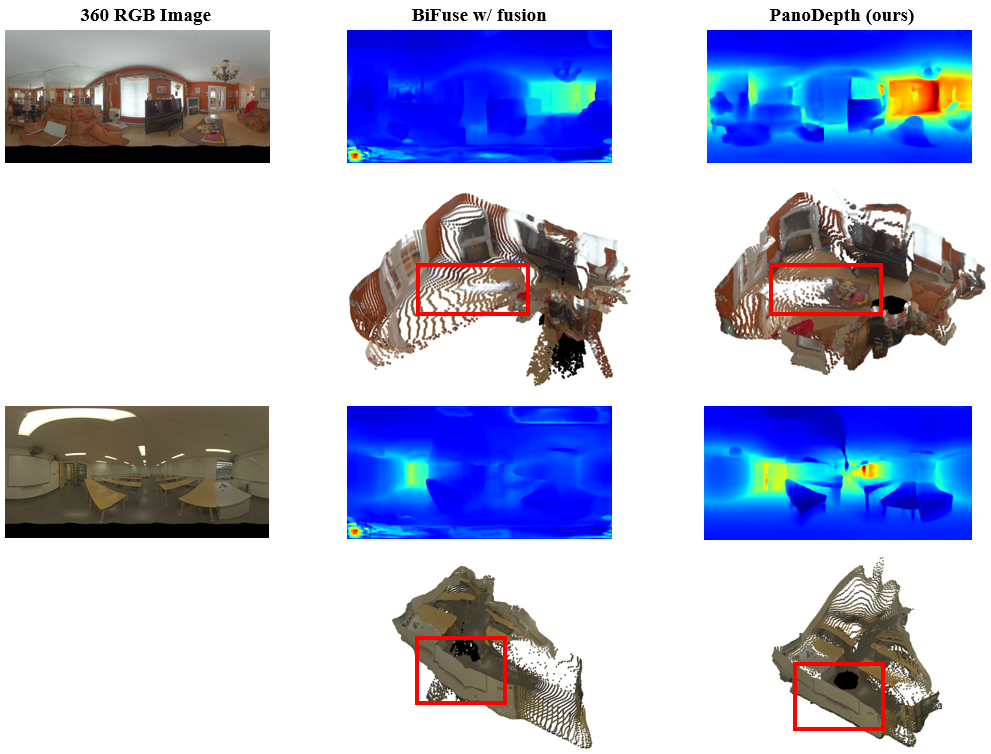}
    \caption{Generalization capability comparison between BiFuse (with fusion) and our PanoDepth. We train the model on 360D \cite{zioulis2018omnidepth} and test on a different real-world dataset, Laval HDR \cite{gardner2017illumination}, which does not have any ground truth depth. Row 1 and 3 are the input images and the estimated depth maps, Row 2 and 4 are the reconstructed point clouds from the two methods. We can see that \textit{PanoDepth} is able to recover the structure of the room, capture the details and sharp edges more accurately.}
    \label{fig:generalization}
\end{figure*}

In Figure \ref{fig:360d}, we show additional qualitative comparisons between our method and the state-of-art approaches \cite{zioulis2018omnidepth,wang2020bifuse} on Stanford2D3d \cite{armeni2017joint}. In Figure \ref{fig:generalization}, we show additional generalization capability comparisons between current state-of-the-art BiFuse (with fusion) and our PanoDepth, with training on 360D \cite{zioulis2018omnidepth} and testing on a different real-world  Laval HDR dataset \cite{gardner2017illumination} with no ground truth depth.
%we test on a real-world 360 dataset that does not have ground truth depth, named the Laval indoor HDR database \cite{gardner2017illumination}. We compare our method with the current state-of-the-art method BiFuse with fusion. 
As can be seen from both figures, our method can more accurately recover the scene structure, predict the depth with less artifacts, and restore the object details, and has stronger generalization ability.

\end{document}